\documentclass[twoside, dvipsnames]{BioRxiv}

\usepackage{epstopdf}
\usepackage{enumitem}
\usepackage{verbatim}
\usepackage{dblfloatfix}
\usepackage{algpseudocode}
\usepackage{amsmath,mathtools}

\usepackage{booktabs,longtable}
\usepackage{graphicx}
\usepackage{dcolumn}
\usepackage{bm}

\usepackage[most]{tcolorbox}

\usepackage{tikz}
\usepackage{pgfplots}
\pgfplotsset{compat=1.13}
\pgfplotsset{
    table/search path={figs/data},
}

\usepackage{xfrac}
\usetikzlibrary{arrows}

\usetikzlibrary{math}
\usetikzlibrary{shapes,arrows}
\usetikzlibrary{positioning}
\usetikzlibrary{decorations,patterns}
\usetikzlibrary{calc,fit,matrix,chains,scopes}

\usepgfplotslibrary{colorbrewer}

\newcommand{\lw}{1pt}

\definecolor{accent1}{HTML}{4267B6}
\definecolor{accent2}{HTML}{003462}
\definecolor{accent3}{HTML}{162643}

\definecolor{myblue}{RGB}{31, 119, 180}
\definecolor{myred}{RGB}{44, 160, 44}
\definecolor{mygreen}{RGB}{255, 127, 14}
\definecolor{myviolet}{RGB}{140, 86, 75}

\tikzset{%
    block/.style        = {line width=\lw, rectangle, draw=black,dashed, minimum width=\dist, minimum height=.33*\dist},
    representation/.style        = {line width=\lw, rectangle, draw=white!0,fill=mygreen, minimum width=.05*\dist, minimum height=.3\dist},
    filled/.style        = {line width=\lw, rectangle, draw=white!0, rounded corners, minimum width=.15*\dist, minimum height=.05*\dist},
    line/.style         = {draw=gray, -latex, line width=.75*\lw,rounded corners},
    lineplain/.style    = {draw=gray, -, line width=.75*\lw, rounded corners},
    mathop/.style       = {draw, circle, fill=gray!20},
    branch/.style       = {fill,shape=circle,minimum size=\dist,inner sep=0pt},
    encoder/.style      = {line width=\lw, trapezium, trapezium angle=80, rotate=-90, minimum width=.55*\dist,  draw, trapezium stretches=true, fill=myblue,draw=white!0,minimum height=.15*\dist},
    classifier/.style      = {line width=\lw, trapezium, trapezium angle=80, rotate=-90, minimum width=.35*\dist, minimum height=.15*\dist, draw, trapezium stretches=true, draw=white},
    decoder/.style      = {line width=\lw, trapezium, trapezium angle=80, rotate=90,  minimum width=.55*\dist, minimum height=.15*\dist, draw=white, trapezium stretches=true},
    mult/.style={path picture={
      \draw[black]
    (path picture bounding box.south east) -- (path picture bounding box.north west) (path picture bounding box.south west) -- (path picture bounding box.north east);
    }},
    add/.style={path picture={
      \draw[black]
    (path picture bounding box.south) -- (path picture bounding box.north) (path picture bounding box.west) -- (path picture bounding box.east);
    }}
}

\makeatletter
\pgfplotsset{
  tufte axes/.style =
    {
      after end axis/.code =
        {
          \draw ({rel axis cs:0,0} -| {axis cs:\pgfplots@data@xmin,0})
            -- ({rel axis cs:0,0}  -| {axis cs:\pgfplots@data@xmax,0});
          \draw ({rel axis cs:0,0} |- {axis cs:0,\pgfplots@data@ymin})
            -- ({rel axis cs:0,0}  |-{axis cs:0,\pgfplots@data@ymax});
        },
      axis line style = {draw = none},
      tick align      = outside,
      tick pos        = left
    }
}
\makeatother

\pgfkeys{/pgfplots/tuftelike/.style={
  thick,
  tick style={major tick length=3pt, thick, black},
  axis x line*=bottom,
  axis line shift = 10pt,
  tick align      = outside,
  tick pos        = left,
  xlabel shift=0pt,
  axis y line*=left,
  ylabel shift=0pt}
}

\newcommand{\dist}{10em}

\pgfplotsset{
    discard if not/.style 2 args={
        x filter/.code={
            \edef\tempa{\thisrow{#1}}
            \edef\tempb{#2}
            \ifx\tempa\tempb
            \else
                
            \fi
        }
    }
}

\newtcolorbox[auto counter]{pabox}[2][]{%
colback=blue!5!white,colframe=blue!75!black,fonttitle=\bfseries,
title=Box~\thetcbcounter: #2,#1}

\shorttitle{Motion Capture with Deep Learning}

\author[1,2,3,5,*]{Alexander Mathis}
\author[3,4]{Steffen Schneider}
\author[1,2,3]{Jessy Lauer}
\author[1,2,3,*]{Mackenzie W. Mathis}

\affil[1]{Center for Neuroprosthetics, Center for Intelligent Systems, Swiss Federal Institute of Technology (EPFL), Lausanne, Switzerland}
\affil[2]{Brain Mind Institute, School of Life Sciences, Swiss Federal Institute of Technology (EPFL), Lausanne, Switzerland}
\affil[3]{The Rowland Institute at Harvard, Harvard University, Cambridge, MA USA}
\affil[4]{University of Tübingen \& International Max Planck Research School for Intelligent Systems, Germany}
\affil[5]{lead Contact}
\affil[*]{Corresponding authors: alexander.mathis@epfl.ch, mackenzie.mathis@epfl.ch}
\leadauthor{Mathis}
\title{A Primer on Motion Capture with Deep Learning: Principles, Pitfalls and Perspectives}

\begin{document}

\maketitle

\begin{abstract}
{\bf Extracting behavioral measurements non-invasively from video is stymied by the fact that it is a hard computational problem. Recent advances in deep learning have tremendously advanced predicting posture from videos directly, which quickly impacted neuroscience and biology more broadly. In this primer we review the budding field of motion capture with deep learning. In particular, we will discuss the principles of those novel algorithms, highlight their potential as well as pitfalls for experimentalists, and provide a glimpse into the future.}
\end{abstract}

\section*{Introduction}
The pursuit of methods to robustly and accurately measure animal behavior is at least as old as the scientific study of behavior itself~\cite{klette2008understanding}. Trails of hominid footprints, ``motion'' captured by pliocene deposits at Laetoli that date to 3.66 million years ago, firmly established that early hominoids achieved an upright, bipedal and free-striding gait~\cite{leakey1979pliocene}. 
Beyond fossilized locomotion, behavior can now be measured in a myriad of ways: from GPS trackers, videography, to microphones, to tailored electronic sensors~\cite{kays2015terrestrial, brown2013observing, camomilla2018trends}. 
Videography is perhaps the most general and widely-used method as it allows noninvasive, high-resolution observations of behavior~\cite{johansson1973visual, o2010camera, weinstein2018computer}. Extracting behavioral measures from video poses a challenging computational problem. Recent advances in deep learning have tremendously simplified this process~\cite{wu2020recent, mathis2020deep}, which quickly impacted neuroscience~\cite{mathis2020deep, datta2019computational}. 
\medskip

In this primer we review markerless (animal) motion capture with deep learning. In particular, we review principles of algorithms, highlight their potential, as well as discuss pitfalls for experimentalists, and compare them to alternative methods (inertial sensors, markers, etc.). Throughout, we also provide glossaries of relevant terms from deep learning and hardware. Furthermore, we will discuss how to use them, what pitfalls to avoid, and provide perspectives on what we believe will and should happen next. 
\medskip

\begin{figure*}[htp]
    \centering
    \includegraphics[width=\textwidth]{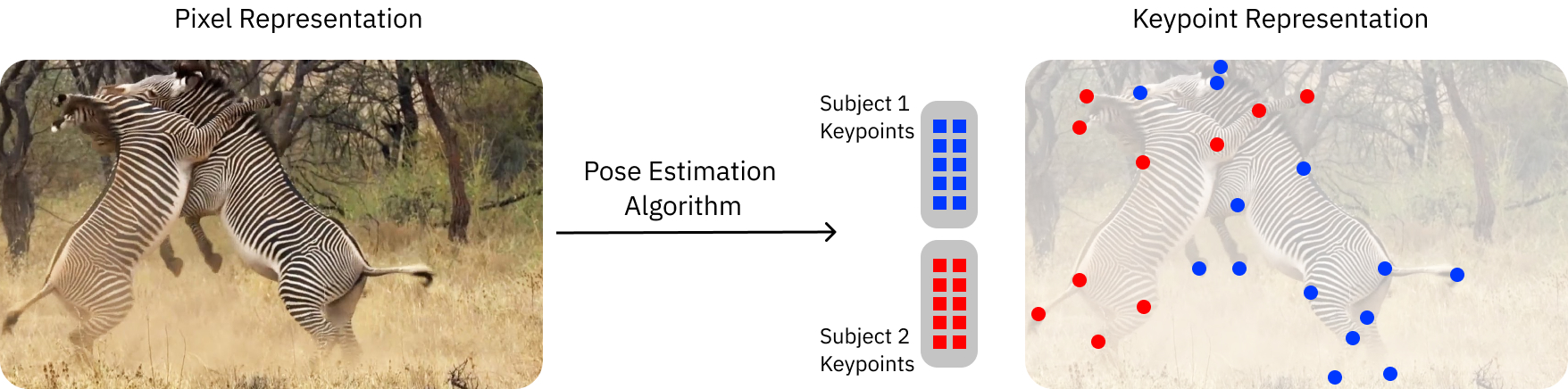}
    \caption{%
    {\bf Schematic overview of markerless motion capture or pose estimation.} The pixel representation of an image (left) or sequence of images (video) is processed and converted into a list of keypoints (right).
    Semantic information about object identity and keypoint type is associated to the predictions. For instance, the keypoints are structures with a name e.g. ear, the x and y coordinate as well as a confidence readout of the network (often this is included, but not for all pose estimation packages), and are then grouped according to individuals (subjects).}
    \label{fig:overview}
\end{figure*}

What do we mean by ``markerless motion capture?'' While biological movement can also be captured by dense, or surface models~\cite{mathis2020deep, guler2018densepose, Zuffi20163dMenagerie}, here we will almost exclusively focus on ``keypoint-based pose estimation.'' Human and many other animal's motion is determined by the geometric structures formed by several pendulum-like motions of the extremities relative to a joint~\cite{johansson1973visual}. Seminal psychophysics studies by Johansson showed that just a few coherently moving keypoints are sufficient to be perceived as human motion~\cite{johansson1973visual}. This empirically highlights why pose estimation is a great summary of such video data. Which keypoints should be extracted, of course, dramatically depends on the model organism and the goal of the study; e.g., many are required for dense, 3D models~\cite{guler2018densepose, sanakoyeu2020transferring, Zuffi20163dMenagerie}, while a single point can suffice for analyzing some behaviors~\cite{mathis2020deep}. One of the great advantages of deep learning based methods is that they are very flexible, and the user can define what should be tracked.

\section*{Principles of deep learning methods for markerless motion capture}

In raw video we acquire a collection of pixels that are static in their location and have varying value over time.
For analyzing behavior, this representation is sub-optimal:
Instead, we are interested in properties of objects in the images, such as location, scale and orientation.
Objects are collections of pixels in the video moving or being changed in conjunction.
By decomposing objects into \emph{keypoints} with semantic meaning ---such as body parts in videos of human or animal subjects---a high dimensional video signal can be converted into a collection of time series describing the movement of each keypoint (Figure~\ref{fig:overview}).
Compared to raw video, this representation is easy to analyze, and semantically meaningful for investigating behavior and addressing the original research question for which the data has been recorded.
\medskip

\begin{figure*}[b]
\centering
\includegraphics[width=\textwidth]{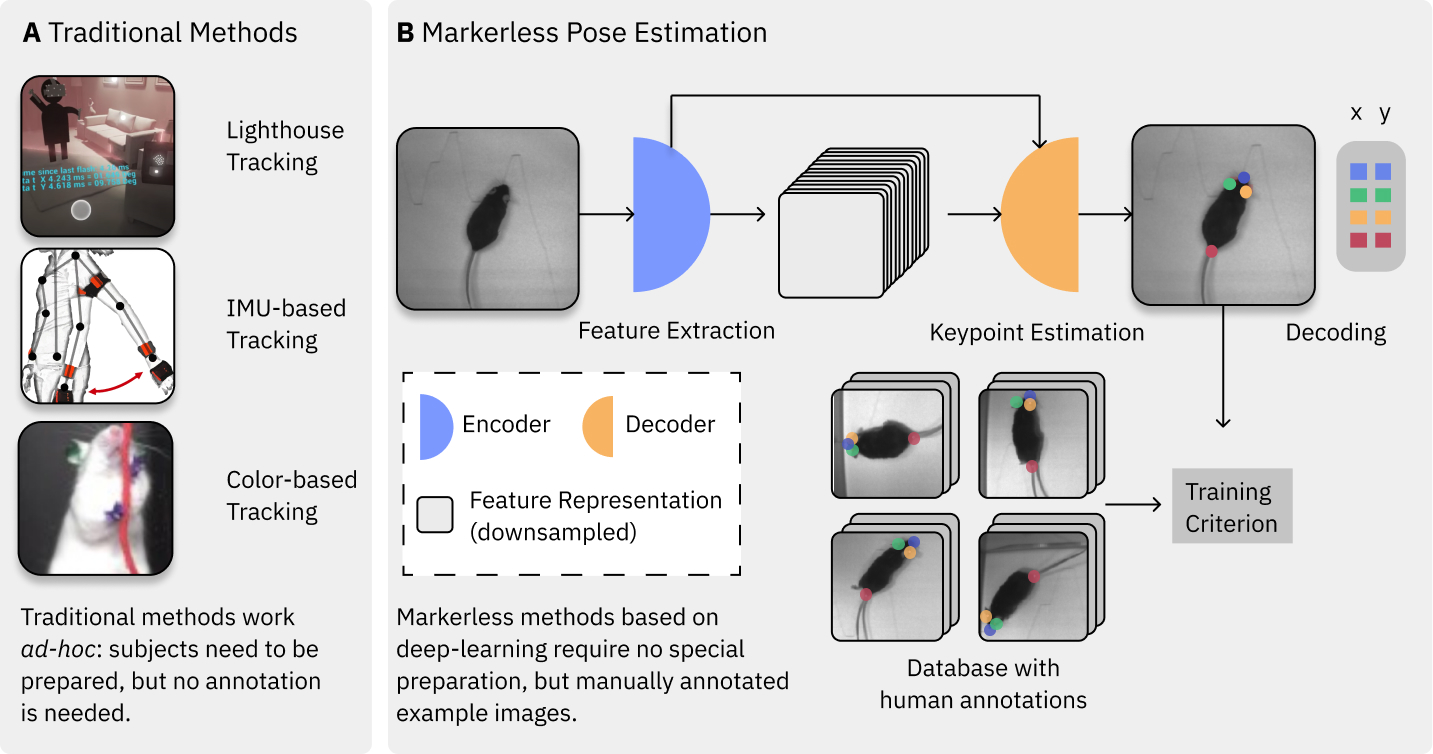}
\caption{{\bf Comparison of marker-based (traditional) and markerless tracking approaches.}
{\bf (A)} In marker-based tracking, \emph{prior} to performing an experiment, special measures have to be taken regarding hardware and preparation of the subject (images adapted from~\citealp{inayat2020matlab, maceira2019wearable}; IMU stands for
inertial measurement unit).
{\bf (B)} For markerless pose estimation, raw video is acquired and processed post-hoc: Using labels from human annotators, machine learning models are trained to infer keypoint representations directly from video (on-line inference without markers is also possible~\cite{Kane2020dlclive}).
Typically, the architectures underlying pose estimation can be divided into a feature extractor and a decoder: The former maps the image representation into a feature space, the latter infers keypoint locations given this feature representation.
In modern deep learning systems, both parts of the systems are trained end-to-end. \label{fig:comparisonmarkerbasedvsmarkerless}
}
\end{figure*}

Motion capture systems aim to infer keypoints from videos:
In marker-based systems, this can be achieved by manually enhancing parts of interest (by colors, LEDs, reflective markers), which greatly simplifies the computer vision challenge, and then using classical computer vision tools to extract these keypoints. Markerless pose estimation algorithms directly map raw video input to these coordinates. The conceptual difference between marker-based and marker-less approaches is that the former requires special preparation or equipment, while the latter can even be applied \emph{post-hoc}, but typically requires ground truth annotations of example images (i.e., a training set). Notably, markerless methods allow for extracting additional keypoints at a later stage, something that is not possible with markers (Figure~\ref{fig:comparisonmarkerbasedvsmarkerless}). 
\medskip

Fundamentally, a pose estimation algorithm can be viewed as a function that maps frames from a video into the coordinates of body parts. The algorithms are highly flexible with regard to what body parts are tracked. Typically the identity of the body parts (or objects) have semantically defined meaning (e.g., different finger knuckles, the head), and the algorithms can group them accordingly (namely, to assemble an individual) so that the posture of multiple individuals can be extracted simultaneously (Figure~\ref{fig:overview}).
For instance, for an image of one human the algorithm would return a list of pixel coordinates (these can have subpixel resolution) per body part and frame (and sometimes an uncertainty prediction;~\citealp{insafutdinov2016deepercut, kreiss2019pifpaf, mathis2018deeplabcut}).
Which body parts the algorithm returns depends on both the application and the training data provided---this is an important aspect with respect to how the algorithms can be customized for applications.

\begin{figure*}[b]
\centering
\includegraphics[width=\textwidth]{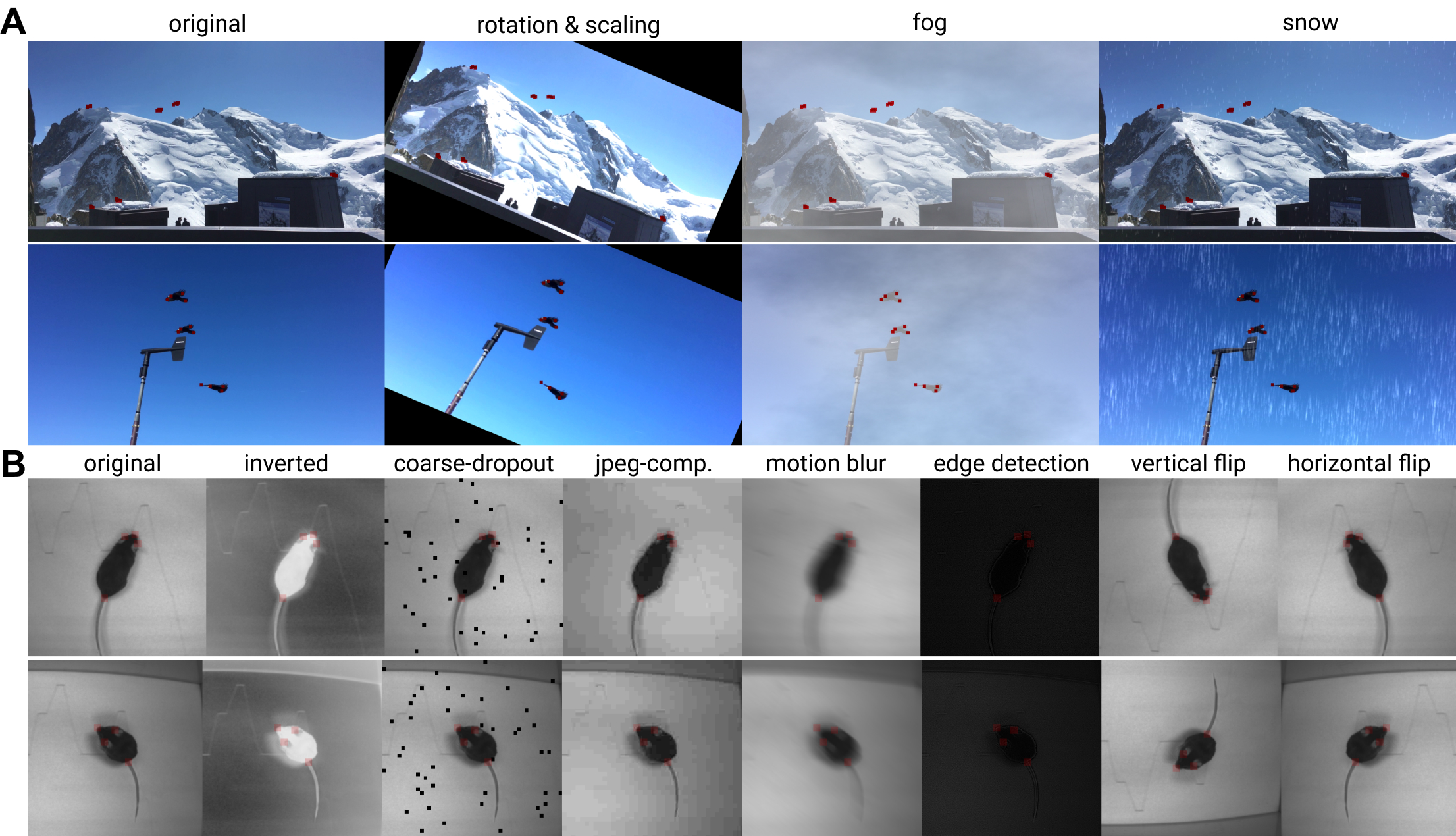}
\caption{Example augmentation images with labeled body parts in red. {\bf (A)} Two example frames of Alpine choughs (Pyrrhocorax graculus) near Mont Blanc with human-applied labels in red (original). The images to the right illustrate three augmentations (as labeled). {\bf (B)} Two example frames of a trail-tracking mouse (mus musculus) from~\cite{mathis2018deeplabcut} with four labeled bodyparts as well as augmented variants. \href{https://colab.research.google.com/github/DeepLabCut/Primer-MotionCapture/blob/master/COLAB_Primer_MotionCapture_Fig3.ipynb}{Open in Google Colaboratory}
}
\label{fig:AUG}
\end{figure*}

\subsection*{Overview of algorithms}
\justify While many pose estimation algorithms~\cite{moeslund2006survey, POPPE20074} have been proposed, algorithms based on deep learning~\citep{lecun2015dl} are the most powerful as measured by performance on human pose estimation benchmarks~\cite{ToshevDEEPPOSE,JainMODEEP,insafutdinov2016deepercut, newell2016stacked, cao2018openpose, Xiao2018, cheng2020higherhrnet}. More generally, pose estimation algorithms fall under ``object detection'', a field that has seen tremendous advances with deep learning (aptly reviewed in Wu et al.,~\citealp{wu2020recent}).
In brief, pose estimation can often intuitively be understood as a system of an encoder that extracts important (visual) features from the frame, which are then used by the decoder to predict the body parts of interests along with their location in the image frame.
\medskip

In classical algorithms (see~\citealp{moeslund2006survey, POPPE20074, wu2020recent}), handcrafted feature representations are used that extract invariant statistical descriptions from images.  These features were then used together with a classifier (decoder) for detecting complex objects like humans~\cite{dalal2005histograms, moeslund2006survey}. Handcrafted feature representations are (loosely) inspired by neurons in the visual pathway and are designed to be robust to changes in illumination, and translations; typical feature representations are Scale Invariant Feature Transform (SIFT; \citealp{lowe2004distinctive}), Histogram of Gradients (HOG; \citealp{dalal2005histograms}) or Speeded Up Robust Features (SURF; ~\citealp{bay2008speeded}).
\medskip

In more recent approaches, both the encoder and decoders (alternatively called the backbone and output heads, respectively) are deep neural networks (DNN) that are directly optimized on the pose estimation task. An optimal strategy for pose estimation is jointly learning representations of the raw image or video data (encoder) and a predictive model for posture (decoder).
In practice, this is achieved by concatenating multiple layers of differentiable, non-linear transformations and by training such a model as a whole using the back propagation algorithm~\cite{lecun2015dl, goodfellow2016deep, wu2020recent}.
In contrast to classical approaches, DNN based approaches directly optimize the feature representation in a way most suitable for the task at hand (For a glossary of deep learning terms see Box~\ref{box1}).
\medskip

Machine learning systems are composed of a dataset, model, loss function (criterion) and optimization algorithm~\cite{goodfellow2016deep}. The dataset defines the input-output relationships that the model should learn: In pose estimation, a particular pose (output) should be predicted for a particular image (input), see Figures~\ref{fig:overview} \& \ref{fig:comparisonmarkerbasedvsmarkerless}B. The model's parameters (weights) are iteratively updated by the optimizer to minimize the loss function. Thereby the loss function measures the quality of a predicted pose (in comparison to the ground truth data). Choices about these four parts influence the final performance and behavior of the pose-estimation system and we discuss possible design choices in the next sections.

\subsection*{Datasets \& Data Augmentation}

\justify Two kinds of datasets are relevant for training pose estimation systems:
First, one or multiple datasets used for related tasks---such as image recognition---can be used for \emph{pre-training} computer vision models on this task (also known as transfer learning; see Box~\ref{box1}). This dataset is typically considerably larger than the one used for pose estimation. For example, ImageNet~\cite{deng2009imagenet}, sometimes denoted as ImageNet-21K, is a highly influential dataset and a subset was used for the ImageNet Large Scale Visual Recognition Challenge in 2012 (ILSRC-2012;~\citealp{russakovsky2015imagenet}) for object recognition. Full ImageNet contains 14.2 million images from 21K classes, the ILSRC-2012 subset contains 1.2 million images of 1,000 different classes (such as car, chair, etc;~\citealp{russakovsky2015imagenet}). Groups working towards state-of-the-art performance on this benchmark also helped push the field to build better DNNs and openly share code. This dataset has been extensively used for pre-training networks, which we will discuss in the model and optimization section below. 
\medskip

The second highly relevant dataset is the one curated for the task of interest-- Mathis et al.~\cite{mathis2018deeplabcut} empirically demonstrated that the size of this dataset can be comparably small for typical pose estimation cases in the laboratory. Typically, this dataset contains 10--500 images, vs. the standard human pose estimation benchmark datasets, such as MS COCO~\cite{lin2014microsoft} or MPII pose~\cite{andriluka20142d}, which has annotated 40K images (of 26K individuals). This implies that the dataset that is curated is highly influential on the final performance, and great care should be taken to select diverse postures, individuals, and background statistics and labeling the data accurately (discussed below in ``pitfalls''). 

\medskip

In practice, several factors matter: the performance of a fine-tuned model on the task of interest, the amount of images that need to be annotated for fine-tuning the network, and the convergence rate of optimization algorithm---i.e., how many steps of gradient descent are needed to obtain a certain performance. Using a pre-trained network can help with this in several regards:
\citet{he2018rethinking} show that in the case of large training datasets, pre-training typically aids with convergence rates, but not necessarily the final performance.
Despite this evidence that under the right circumstances (i.e., given enough task-relevant data) and with longer training, randomly initialized models can match the performance of fine-tuned ones for key point detection on COCO~\cite{he2018rethinking} and horses~\cite{mathis2019TRANSFER}, however, the networks are less robust~\cite{mathis2019TRANSFER}. Beyond robustness, using a pre-trained model is generally advisable when the amount of labeled data for the target task is small, which is true for many applications in neuroscience, as it leads to shorter training times and better performance with less data~\cite{he2018rethinking, mathis2018deeplabcut, mathis2019TRANSFER, arac2019deepbehavior}. Thus, pre-trained pose estimation algorithms save training time, increase robustness, and require substantially less training data. Indeed, most packages in Neuroscience now use pre-trained models~\cite{mathis2018deeplabcut,graving2019fast,arac2019deepbehavior,Bala2020, Liu2020optiflex, mathisimagenet2020}, although some do not~\cite{pereira2019fast,Gnel2019DeepFly3D, Zimmermann2020}, which can give acceptable performance for simplified situations with aligned individuals.
\medskip

More recently, larger datasets like the 3.5 billion Instagram dataset~\cite{mahajan2018exploring}, JFT which has 300M images~\cite{hinton2015distilling,xie2020noisy} and OpenImages~\cite{kuznetsova2018open} became popular, further improving performance and robustness of the considered models~\cite{xie2020noisy}. What task is used for pre-training also matters. Corroborating this insight, Li et al. showed that pre-training on large-scale object detection task can improve performance for tasks that require fine, spatial information like segmentation~\cite{li2019analysis}.
\medskip

Besides large datasets for pre-training, a curated dataset with pose annotations is needed for optimizing the algorithm on the pose estimation task. The process is discussed in more detail below and it typically suffices to label a few (diverse) frames. Data augmentation is the process of expanding the training set by applying specified manipulations (like rotate, scaling image size).
Based on the chosen corruptions, models become more invariant to rotations, scale changes or translations and thus more accurate (with less training data).
Augmentation can also help with improving robustness to noise, like jpeg compression artefacts and motion blur (Figure~\ref{fig:AUG}).
To note, data augmentation schemes should not affect the semantic information in the image: for instance, if color conveys important information about the identity of an animal, augmentations involving changes in color are not advisable.
Likewise, augmentations which change the spatial position of objects or subjects should always be applied to both the input image and the labels (Box~\ref{box2}).

\subsection*{Model architectures}

\justify Systems for markerless pose estimation are typically composed of a \emph{backbone} network (encoder), which takes the role of the feature extractor, and one or multiple \emph{heads} (decoders). Understanding the model architectures and design choices common in deep learning based pose estimation systems requires basic understanding of convolutional neural networks. We summarize the key terms in Box~\ref{box1}, and expand on what encoders and decoders are below.
\medskip

Instead of using handcrafted features as in classical systems, deep learning based systems employ ``generic'' encoder architectures  which are often based on models for object recognition. In a typical system, the encoder design affects the most important properties of the algorithms such as its inference speed, training-data requirements and memory demands. For the pose estimation algorithms so far used in neuroscience the encoders are either stacked hourglass networks~\cite{newell2016stacked}, MobileNetV2s~\cite{sandler2018mobilenetv2}, ResNets~\cite{He_2016_CVPR}, DenseNets~\cite{huang2017densely} or EfficientNets~\cite{tan2019efficientnet}.
These encoder networks are typically pre-trained on one or multiple of the larger-scale datasets introduced previously (such as ImageNet), as this has been shown to be an advantage for pose estimation on small lab-scale sized datasets~\cite{mathis2019TRANSFER, mathis2018deeplabcut, arac2019deepbehavior}.
For common architectures this pre-training step does not need to be carried out explicitly-trained weights for popular architectures are already available in common deep learning frameworks.
\medskip

The impact of the encoder on DNN performance is a highly active research area. The encoders are continuously improved in regards to speed and object recognition performance~\cite{huang2017densely, sandler2018mobilenetv2, tan2019efficientnet, wu2020recent, kornblith2019better}.
Naturally, due to the importance of the ImageNet benchmark the accuracy of network architectures continuously increases (on that dataset). 
For example, we were able to show that this performance increase is not merely reserved for ImageNet, or (importantly) other object recognition tasks~\cite{kornblith2019better}, but in fact that better architectures on ImageNet are also better for pose estimation~\cite{mathisimagenet2020}. However, being better on ImageNet, also comes at the cost of decreasing inference speed and increased memory demands.  DeepLabCut (an open source toolbox for markerless pose estimation popular in neuroscience) thus incorporates backbones from MobileNetV2s (faster) to EfficientNets (best performance on ImageNet; \citealp{mathis2019TRANSFER,mathisimagenet2020}). 
\medskip

{%
\onecolumn
\begin{pabox}[label={box1},nameref={Hardware},size=small, breakable, float, floatplacement=t]{Glossary of Deep Learning terms}

An excellent textbook for Deep Learning is provided by Goodfellow et al.~\cite{goodfellow2016deep}.
See Dumoulin \& Visin \cite{dumoulin2016guide} for an in-depth technical overview of convolution arithmetic.\\

\newcommand{\tblgraphic}[1]{%
\includegraphics[width=.15\textwidth]{#1}
}

\newcommand{\wraptablfig}[1]{
\raisebox{-\height}{\resizebox{.1\columnwidth}{!}{#1}}
}

\small
\begin{tabular}[t]{p{.1\columnwidth}p{.3\columnwidth}p{.1\columnwidth}p{.3\columnwidth}}
\wraptablfig{%
    \begin{tikzpicture}
    \foreach \ya in {0,...,4} {
    \foreach \yb in {0,...,3} {
        \draw  (0,\ya) -- (1.5,.5+\yb);
        \draw  (3,\ya) -- (1.5,.5+\yb);
    }}
    \foreach \yoff in {0,...,4} {
        \fill [fill=blue!40] (0,\yoff) circle (.3);
    }
    \foreach \yoff in {0,...,3} {
        \fill [fill=blue!40] (1.5,.5+\yoff) circle (.3);
    }
    \foreach \yoff in {0,...,4} {
        \fill [fill=blue!40] (3,\yoff) circle (.3);
    }
    \end{tikzpicture}
}
& \textit{{\bf Artificial neural network (ANN):}}
An ANN can be represented by a collection of computational units (``neurons'') arranged in a directed graph. 
The output of each unit is computed as a weighted sum of its inputs, followed by a nonlinear function.
& 
\wraptablfig{%
\wraptablfig{\tblgraphic{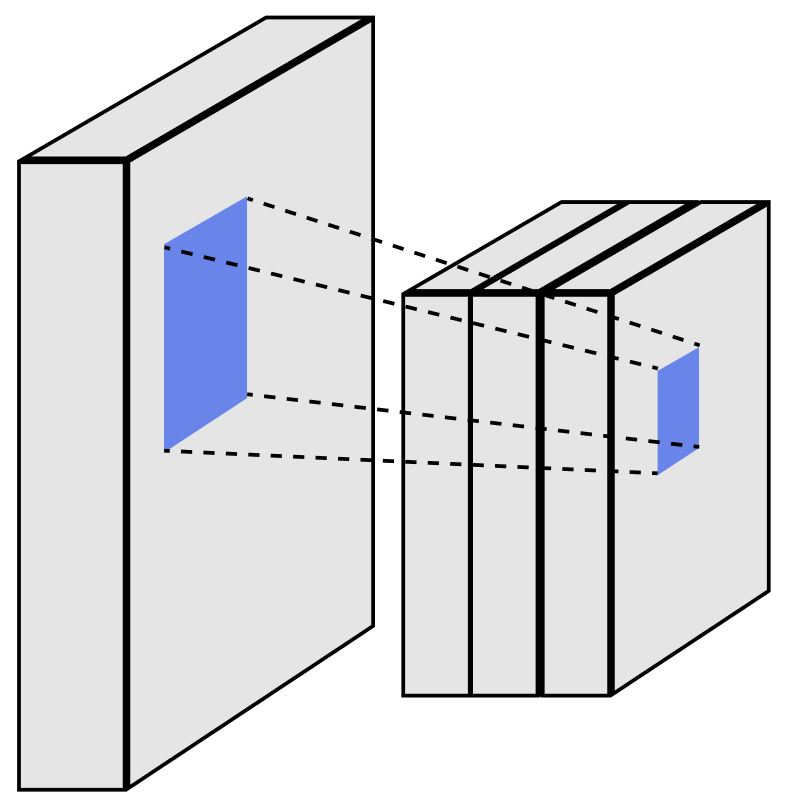}}
}
& \textit{{\bf Convolutional neural network (CNN):}}
 A CNN is an ANN composed of one or multiple convolutional layers.
 Influential early CNNs are the LeNet, AlexNet and VGG16~\cite{lecun2015dl,goodfellow2016deep, wu2020recent}.
 \\[1em]
\wraptablfig{\tblgraphic{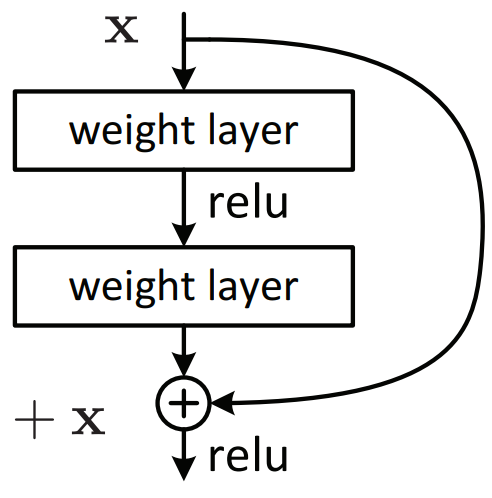}}
 & \multicolumn{3}{p{.75\columnwidth}}{\textit{ {\bf Residual Networks (ResNets):}}
 Increasing network depth makes deep neural networks (DNNs) more expressive compared to adding units to a shallow architecture. However, optimization becomes hard for standard CNNs beyond 20 layers, at which point depth in fact decreases the performance~\cite{He_2016_CVPR}.
 In residual networks, instead of learning a mapping $y = f(x)$, the layer is re-parametrized to learn the mapping $y = x + f(x)$, which improves optimization and regularizes the loss landscape \cite{li2018visualizing}.
 These networks can have much larger depth without diminishing returns~\cite{He_2016_CVPR} and are the basis for other popular architectures such as MobileNets~\citep{sandler2018mobilenetv2} and EfficientNets~\cite{tan2019efficientnet}.}
\end{tabular}

\begin{tabular}[t]{p{.1\columnwidth}p{.75\columnwidth}}
 \wraptablfig{\tblgraphic{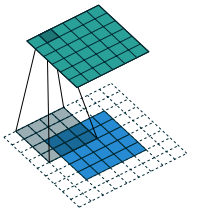}}
 &
 \textit{ {\bf Convolution:}}
  A convolution is a special type of linear filter.
  Compared with a full linear transformation, convolutional layers increase computational efficiency by weight sharing~\cite{lecun2015dl,goodfellow2016deep}. By applying the convolution, the same set of weights is used across all locations in the image.
 \textit{{\bf Deconvolution:}}
 Deconvolutional layers allow to upsample a feature representation.
 Typically, the kernel used for upsampling is optimized during training, similar to a standard convolutional layer.
 Sometimes, fixed operations such as bilinear upsampling filters are used.
 \\[1em]
\wraptablfig{\tblgraphic{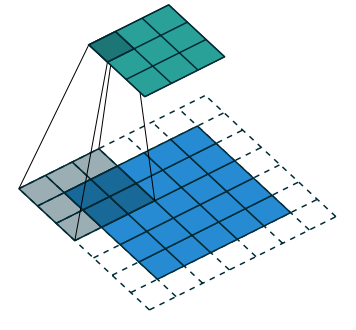}}
 & \textit{ {\bf Stride, Downsampling and Dilated (atrous) Convolutions:}} 
 In DNNs for computer vision, images are presented as real-valued pixel data to the network and are then transformed to symbolic representations and image annotations, such as bounding boxes, segmentation masks, class labels or keypoints.
 During processing, inputs are consecutively abstracted by aggregating information from growing ``receptive fields''.
 Increasing the receptive field of the unit is possible by different means:
 Increasing the stride of a layer computes outputs only for every $n$-th input and effectively downsamples the input with a learnable filter.
 Downsampling layers perform the same operation, but with a fixed kernel (e.g. taking the maximum or mean activation across the receptive field).
 In contrast, \emph{atrous} or dilated convolutions increase the filter size by adding intermittent zero entries between the learnable filter weights---e.g., for a dilation of 2, a filter with entries $(1,2,3)$ would be converted into $(1,0,0,2,0,0,3)$.
 This allows increases in the receptive field without loosing resolution in the next layers, and is often applied in semantic segmentation algorithms~\cite{chen2017deeplab}, and pose estimation~\cite{insafutdinov2016deepercut,mathis2018deeplabcut}.
 \\
 \raisebox{-.65\height}{\resizebox{.1\columnwidth}{!}{\tblgraphic{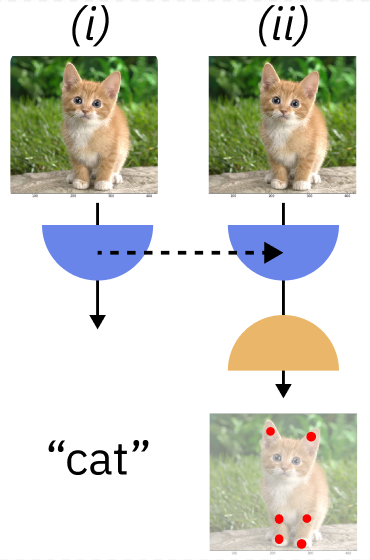}}}
& 
{\textit{ {\bf Transfer learning:}}
The ability to use parameters from a network that has been trained on one task---e.g. classification, see \emph{(i)}---as part of a network to perform another task---e.g. pose estimation, see \emph{(ii)}.
The approach was popularized with DeCAF~\cite{donahue2014decaf}, which used AlexNet \cite{krizhevsky2012imagenet} to extract features to achieve excellent results for several computer vision tasks. Transfer learning generally improves the convergence speed of model training \cite{he2018rethinking, zamir2018taskonomy, arac2019deepbehavior, mathis2019TRANSFER} and model robustness compared to training from scratch \cite{mathis2019TRANSFER}.
}
\end{tabular}

\end{pabox}
}
\twocolumn

\begin{pabox}[label={box2},nameref={Hardware}]{Key parameters and choices.}
The key design choices of pose estimation systems are dataset curation, data augmentation, model architecture selection, optimization process, and the optimization criterions.
\begin{itemize}[leftmargin=*]

\item \textit{{\bf Data Augmentation}}: is the technique of increasing the training set by converting images and annotations into new, altered images via geometric transformations (e.g. rotation, scaling..),  image manipulations (e.g. contrast, brightness,...), etc. (Figure~\ref{fig:AUG}) . Depending on the annotation data, various augmentations, i.e. rotation symmetry/etc. are ideal. Packages such as Tensorpack~\cite{wu2016tensorpack} and imgaug~\cite{imgaug} as well as tools native to PyTorch~\cite{paszke2019pytorch} and TensorFlow~\cite{abadi2016tensorflow} provide common augmentation methods and are used in many packages.
\item \textit{{\bf Model Architecture}}:
Users should select an architecture that is accurate and fast (enough) for their goal. Top performing networks (in terms of accuracy) include Stacked Hourglass~\cite{newell2016stacked}, ResNets~\cite{He_2016_CVPR} and EfficientNets~\cite{tan2019efficientnet} with appropriate decoders~\cite{insafutdinov2016deepercut, Xiao2018, kreiss2019pifpaf, mathisimagenet2020} as well as recent high-resolution nets~\cite{sun2019deep, cheng2020higherhrnet}. Performance gains in speed at the expense of slightly worse accuracy are possible with (optimized) lightweight models such as MobileNetV2~\cite{sandler2018mobilenetv2} in DeepLabCut~\cite{mathis2019TRANSFER} and stacked hourglass networks with DenseNets~\cite{huang2017densely} as proposed in DeepPoseKit~\cite{graving2019fast}; and often this performance gap can be rescued with good data augmentation.
\end{itemize}
\end{pabox}

\begin{figure}[b]
\centering
\includegraphics[width=0.5\textwidth]{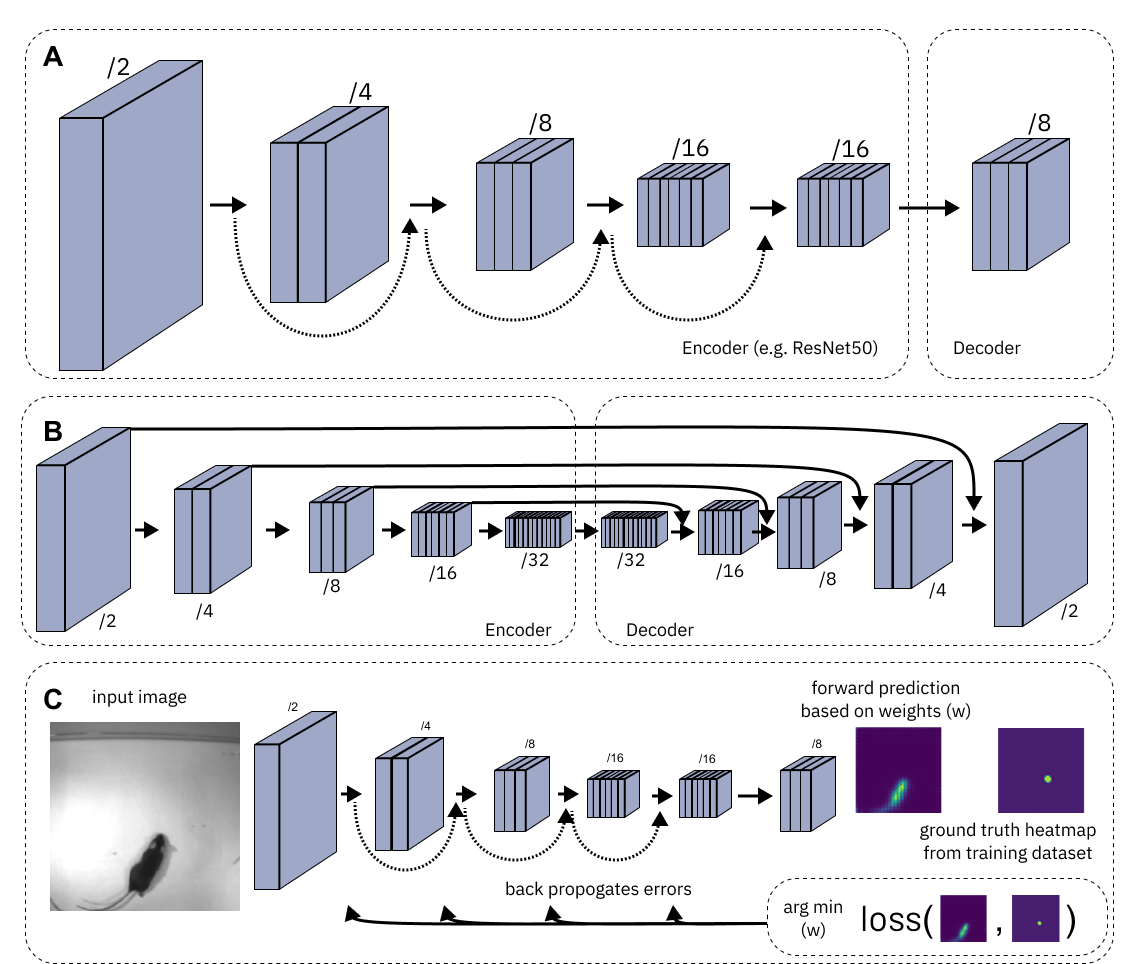}
\caption{%
{\bf Schematic overview of possible design choices for model architectures and training process} {\bf(A)} A simple, but powerful variant~\cite{insafutdinov2016deepercut} is a ResNet-50~\cite{He_2016_CVPR} architecture adapted to replace the final down-sampling operations by atrous convolutions~\cite{chen2017deeplab} to keep a stride of 16, and then a single deconvolution layer to upsample to output maps with stride 8. It also forms the basis of other architectures (e.g.~\citealp{Xiao2018}). The encoder can also be exchanged for different backbones to improve speed or accuracy (see Box~\ref{box2}).
{\bf(B)} Other approaches like stacked hourglass networks~\cite{newell2016stacked}, are not pre-trained and employ skip connections between encoder and decoder layers to aid the up-sampling process. {\bf(C)} For training the network, the training data comprising input images and target heatmaps is used. The target heatmap is compared with the forward prediction. Thereby, the parameters of the network are optimized to minimize the loss that measures the difference between the predicted heatmap and the target heatmap (ground truth). 
}
\label{fig:model-architectures}
\end{figure}

In (standard) convolutional encoders, the high-resolution input images get gradually downsampled while the number of learned features increases. Regression based approaches which directly predict keypoint locations from the feature representation can potentially deal with this downsampled representation. When the learning problem is instead cast as identifying the keypoint locations on a grid of pixels, the output resolution needs to be increased first, often by deconvolutional layers~\cite{insafutdinov2016deepercut, Xiao2018}. We denote this part of the network as the decoder, which takes downsampled features, possibly from multiple layers in the encoder hierarchy, and gradually upsamples them again to arrive at the desired resolution. The first models of this class were Fully Convolutional Networks~\cite{long2015fully}, and later DeepLab~\cite{chen2017deeplab}. Many popular architectures today follow similar principles. Design choices include the use of skip connections between decoder layers, but also regarding skip connections between the encoder and decoder layers. Example encoder--decoder setups are illustrated in Figure~\ref{fig:model-architectures}. The aforementioned building blocks---encoders and decoders---can be used to form a variety of different approaches, which can be trained end-to-end directly on the target task (i.e., pose estimation).
\medskip

Pre-trained models can also be adapted to a particular application. For instance, DeeperCut~\cite{insafutdinov2016deepercut}, which was adapted by the animal pose estimation toolbox DeepLabCut~\cite{mathis2018deeplabcut}, was built with a ResNet~\cite{He_2016_CVPR} backbone network, but adapted the stride by atrous convolutions~\cite{chen2017deeplab} to retain a higher spatial resolution (Box~\ref{box1}). This allowed larger receptive fields for predictions, but retains a relatively high speed (i.e., for video analysis) but most importantly because ResNets can be pre-trained on ImageNet, those initialized weights could be used. Other architectures, like stacked hourglass networks~\cite{newell2016stacked} used in DeepFly3D~\cite{pavan2019} and DeepPoseKit~\cite{graving2019fast}, retain feature representations at multiple scales and pass those to the decoder (Figure~\ref{fig:model-architectures}A, B).

\begin{figure*}[b]
    \centering
    \includegraphics[width=\textwidth]{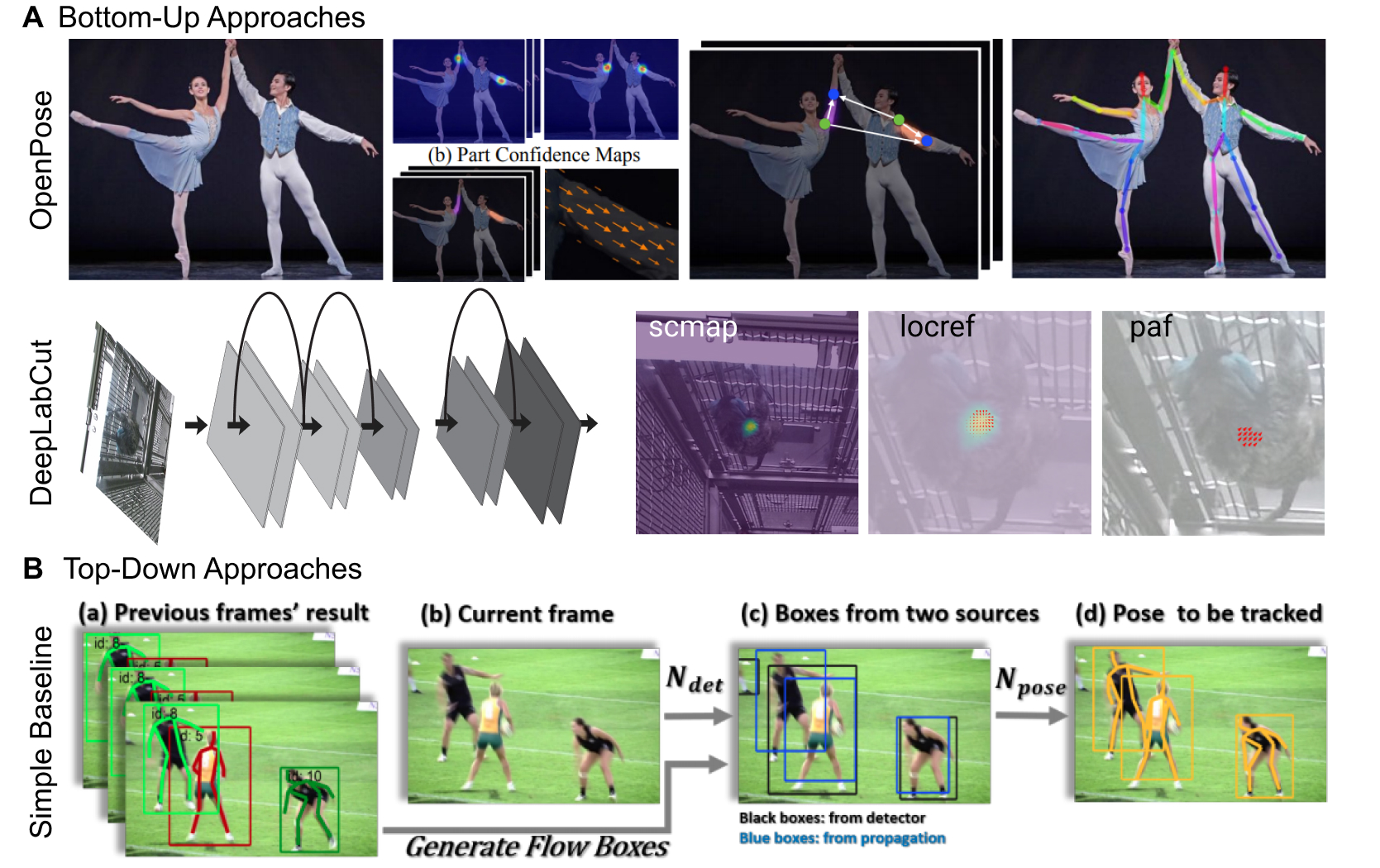}
    \caption{%
    {\bf Multi-animal pose estimation approaches.} {\bf A}: Bottom-up approaches detect all the body parts (e.g. elbow and shoulder in example) as well as ``limbs'' (part confidence maps). These limbs are then used to associate the bodyparts within individuals correctly (Figure from OpenPose,~\citealp{cao2018openpose}). For both OpenPose and DeepLabCut, the bodyparts and part confidence maps, and part affinity fields (paf's) are predicted as different decoders (aka output heads) from the encoder. 
    {\bf B}: Top-down approaches localize individuals with bounding-box detectors and then directly predict the posture within each bounding box. This does not require part confidence maps, but is subject to errors when bounding boxes are wrongly predicted (see black bounding box encompassing two players in (c)). The displayed figures, adapted from Xiao et al.~\cite{Xiao2018}, improved this disadvantage by predicting bounding boxes per frame and forward predicting them across time via visual flow.}
    \label{fig:bottom-up_top-down}
\end{figure*}

\subsection*{Loss functions: training architectures on datasets}

\justify Keypoints (i.e., bodyparts) are simply coordinates in image space. There are two fundamentally different ways for estimating keypoints (i.e., how to define the loss function). The problem can be treated as a regression problem with the coordinates as targets~\cite{ToshevDEEPPOSE, carreira2016human}. Alternatively, and more popular, the problem can be cast as a classification problem, where the coordinates are mapped onto a grid (e.g. of the same size as the image) and the model predicts a heatmap (scoremap) of location probabilities for each bodypart (Figure~\ref{fig:model-architectures}C).
In contrast to the regression approach~\cite{ToshevDEEPPOSE}, this is fully convolutional, and allows modeling of multi-modal distributions and aids the training process~\cite{tompson2014joint, newell2016stacked, insafutdinov2016deepercut, cao2018openpose}. Moreover, the heatmaps have the advantage that one can naturally predict multiple locations of the ``same'' bodypart in the same image (i.e., 2 elbows) without mode collapse (Figure~\ref{fig:bottom-up_top-down}A). 
\medskip

Loss functions can also reflect additional priors or inductive biases about the data. For instance, DeepLabCut uses location refinement layers (locref), that counteract the downsampling inherent in encoders, by training outputs to predict corrective shifts in image coordinates relative to the downsampled output maps (Figure~\ref{fig:bottom-up_top-down}A). In pose estimation, it is possible to define a \emph{skeleton} or graph connecting keypoints belonging to subjects with the same identity (see below)~\cite{insafutdinov2016deepercut,cao2018openpose}. When estimating keypoints over time, it is also possible to employ temporal information and encourage the model to only smoothly vary its estimate among consecutive frames~\cite{insafutdinov2017cvpr,yao2019monet, xu2020eventcap,zhou2020monocular}. Based on the problem, these priors can be directly encoded and be used to regularize the model.
\medskip

How can pose estimation algorithms accommodate multiple individuals? Fundamentally, there are two different approaches: bottom-up and top-down methods (Figure~\ref{fig:bottom-up_top-down}). In top-down methods, individuals are first localized (often with another neural network trained on object localization) then pose estimation is performed per localized individual~\cite{Xiao2018,newell2016stacked,sun2019deep}. In bottom-up methods all bodyparts are localized, and networks are also trained to predict connections of bodyparts within individuals (i.e., limbs). These connections are then used to link candidate bodyparts to form individuals~\cite{cao2018openpose, insafutdinov2017cvpr,kreiss2019pifpaf,cheng2020higherhrnet}. To note, these techniques can be used on single individuals for increased performance, but often are not needed and usually imply reduced inference speed.

\subsection*{Optimization}

\justify For pre-training, stochastic gradient descent (SGD; \citealp{bottou2010large}) with momentum~\cite{sutskever2013importance} is an established method.
Different variants of SGD are now common (such as Adam;~\citealp{kingma2014adam}) and used for fine-tuning the resulting representations.
As mentioned above, pose estimation algorithms are typically trained in a multi-stage setup where the backbone is trained first on a large (labeled) dataset of a potentially unrelated task (like image classification). Users can also download these pre-trained weights. Afterwards, the model is fine-tuned on the pose-estimation task. Once trained, the quality of the prediction can be judged in terms of the root mean square error (RMSE), which measures the distance between the ground truth keypoints and predictions~\cite{mathis2018deeplabcut,pereira2019fast}, or by measuring the percentage of correct keypoints (PCK,~\citealp{andriluka20142d, mathis2019TRANSFER}); i.e., the fraction of detected keypoints that fall within a defined distance of the ground truth.
\medskip

To properly estimate model performance in an application setting, it is advisable to split the labeled dataset at least into train and test subsets. 
If systematic deviations can be expected in the application setting (e.g., because the subjects used for training the model differ in appearance from subjects encountered at model deployment~\cite{mathis2019TRANSFER}, this should be reflected when choosing a way to split the data. For instance, if data from multiple individuals is possible, distinct individuals should form distinct subsets of the data.
On the contrary, strategies like splitting data by selecting every \textit{n}-th frame in a video likely overestimates the true model performance.
\medskip

The model is then optimized on the training dataset, while performance is monitored on the validation (test) split.
If needed, hyperparameters---like parameter settings of the optimizer, or also choices about the model architecture---of the model can be adapted based on an additional validation set.

\medskip

\begin{figure*}[t]
\centering
\includegraphics[width=\textwidth]{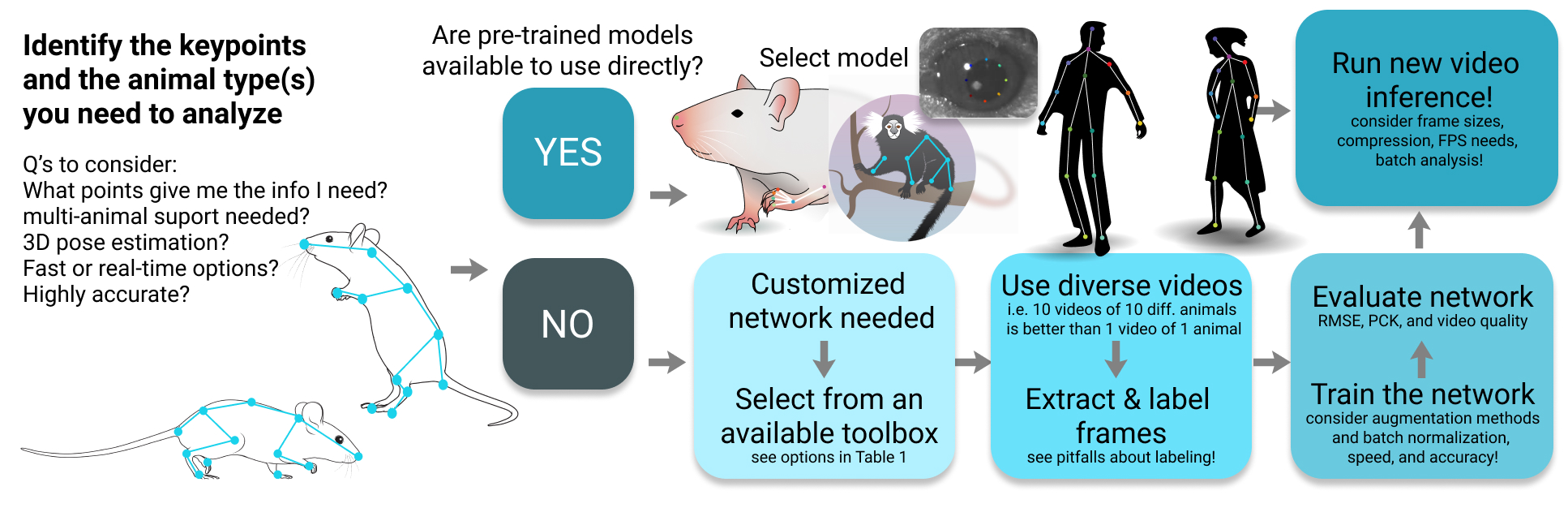}
\caption{An overview of the workflow for deep learning based pose estimation, which highlights several critical decision points.}
\label{fig:workflow}
\end{figure*}
\medskip

\begin{table*}[b]
\small
\caption{Overview of popular deep learning tools for animal motion capture (or newly presented packages that, minimally, include code). Here, we denote if it can be used to create tailored networks, or only specific animal tools are provided, i.e., only work ``as-is'' on a fly or rat. We also only highlight if beyond human pre-trained neural networks (PT-NNs) are available. We also provide the release date and current citations for noted references, including those to related preprints (indexed from google scholar). *note, this code is deprecated and supplanted by SLEAP.}
\begin{tabular}{lccccccccc}
\toprule
 & Any species & 3D & $>$1 animal & Training Code & Full GUI & Ex. Data & PT-NNs & Released & Citations \\
\midrule

DeepLabCut~\cite{mathis2018deeplabcut, nath2019deeplabcut} & yes & yes & yes & yes & yes & yes  & many & 4/2018 & 491\\
LEAP~\cite{pereira2019fast} & yes & no & yes & yes & yes & yes  & no & 6/2018* & 98\\ 
DeepBehavior~\cite{arac2019deepbehavior} & no & yes & yes & no & no & no & no & 5/2019 & 15\\
DeepPoseKit~\cite{graving2019fast} & yes & no & no & yes & partial & yes  & no & 8/2019 & 48\\
DeepFly3D~\cite{pavan2019} & no & yes & no & 2D only & partial & yes  & fly &5/2019 & 21\\ 
FreiPose~\cite{Zimmermann2020} & no & yes & no & partial & no & yes & no & 2/2020 & 1\\
Optiflex~\cite{Liu2020optiflex} & yes & no& no & yes & partial & yes & no & 5/2020 & 0\\
\bottomrule
\end{tabular}
\label{tab:packages}
\end{table*}

All of the aforementioned choices influence the final outcome and performance of the algorithm. While some parts of the training pipeline are well-established and robust---like pre-training a model on ImageNet---choices about the dataset, architecture, augmentation, fine-tuning procedure, etc. will inevitably influence the quality of the pose estimation algorithm (Box~\ref{box2}). See Figure~\ref{fig:AUG} for a qualitative impression of augmentation effects of some of these decisions (see also Figure~\ref{fig:AUG2}). We will discuss this in more detail in the Pitfalls section.
\medskip

So far, we considered algorithms able to infer 2D keypoints from videos, by training deep neural networks on previously labeled data. Naturally, there is also much work in computer vision and machine learning towards the estimation of 3D keypoints from 2D labels, or to directly infer 3D keypoints. In the interest of space, we had to omit those but refer the interested reader to ~\cite{martinez2017simple, Mehta2017_3D, TomeRA17,chen20173d,yao2019monet} as well specifically for neuroscience~\cite{yao2019monet, pavan2019, nath2019deeplabcut, Zimmermann2020, karashchuk2020anipose, Bala2020}. 
\medskip

Lastly, it is not understood how CNNs make decisions and they often find ``shortcuts''~\cite{geirhos2020shortcut}. While this active research area is certainly beyond the scope of this primer, from practical experience we know that at least within-domain---i.e., data that is similar to the training set---DNNs work very well for pose estimation, which is the typical setting relevant for downstream applications in neuroscience. It is worth noting that in order to optimize performance, there is no one-size-fits-all solution. Thus, we hope by building intuition in users of such systems, we provide the necessary tools to make these decisions with more confidence (Figure~\ref{fig:workflow}).

\section*{Scope and applications}

Markerless motion capture can excel in complicated scenes, with diverse animals, and with any camera available (mono-chrome, RGB, depth cameras, etc). The only real requirement is the ability of the human to be able to reliably label keypoints (manually or via alternative sources). Simply, you need to be able to see what you want to track. Historically, due to limitations in computer vision algorithms experimentalists would go to great lengths to simplify the environment, even in the laboratory (i.e., no bedding, white or black walls, high contrast), and this is no longer required with deep learning-based pose estimation. Now, the aesthetics one might want for photographs or videos taken in daily life are the best option. 
\medskip

Indeed, the field has been able to rapidly adopt these tools for neuroscience. Deep learning-based markerless pose estimation applications in the laboratory have already been published for flies~\cite{mathis2018deeplabcut, pereira2019fast, graving2019fast, pavan2019, karashchuk2020anipose,Liu2020optiflex}, rodents~\cite{mathis2018deeplabcut,MathisWarren2018speed, pereira2019fast, graving2019fast, arac2019deepbehavior, pavan2019,Zimmermann2020,Liu2020optiflex}, horses~\cite{mathis2019TRANSFER}, dogs~\cite{yao2019monet}, rhesus macaque~\cite{berger2020wireless, yao2019monet, Bala2020, labuguen2020macaquepose} and marmosets~\cite{ebina2019arm}; the original architectures were developed for humans~\cite{insafutdinov2016deepercut, newell2016stacked, cao2018openpose}. Outside of the laboratory, DeepPoseKit was used for zebras~\cite{graving2019fast} and DeepLabCut for 3D tracking of cheetahs~\cite{nath2019deeplabcut}, for squirrels~\cite{barrett2020manual} and macaques~\cite{labuguen2020macaquepose}, highlighting the great ``in-the-wild'' utility of this new technology~\cite{mathis2020deep}. As outlined in the principles section, and illustrated by these applications, these deep learning architectures are general-purpose and can be broadly applied to any animal as well as condition.
\medskip

Recent research highlights the prevalent representations of action across the brain~\cite{kaplan2020brain}, which emphasizes the importance of quantifying behavior even in non-motor tasks. For instance, pose estimation tools have recently been used to elucidate the neural variability across cortex in humans during thousands of spontaneous reach movements~\cite{peterson2020behavioral}. 
Pupil tracking is of great importance for visual neuroscience. One recent study by Meyer et al. used head-fixed cameras and DeepLabCut to reveal two distinct types of coupling between eye and head movements~\cite{meyer2020two}. In order to accurately correlate neural activity to visual input tracking the gaze is crucial. The recent large, open dataset from the Allen Institute includes imaging data of six cortical and two thalamic regions in response to various stimuli classes as well as pupil tracking with DeepLabCut~\cite{siegle2019survey}. The International Brain Lab has integrated DeepLabCut into their workflow to track multiple bodyparts of decision-making mice including their pupils~\cite{Harris2020dataIBL}.
\medskip

Measuring relational interactions is another major direction, that has been explored less in the literature so far, but is feasible. Since the feature detectors for pose estimation are of general nature one can easily not only track the posture of individuals but also the tools and objects one interacts with (e.g. for analyzing golf or tennis). Furthermore, social behaviors, and parenting interactions (for example in mice) can now be studied noninvasively. 
\medskip

Due to the general capabilities, these tools have several applications for creating biomarkers by extracting high fidelity animal traits, for instance in the pain field~\cite{tracey2019composite} and for monitoring motor function in healthy and  diseased conditions~\cite{micera2020advanced}.

DeepLabCut was also integrated with tools for x-ray analysis~\cite{laurence2020integrating}. For measuring joint center locations in mammals, arguably, x-ray is the gold standard. Of course, x-ray data also poses challenges for extracting body part locations from x-ray data. A recent paper shared methodology to integrate DeepLabCut with XROMM, a popular analysis suite, to advance the speed and accuracy for x-ray based analysis~\cite{laurence2020integrating}.

\section*{How do the (current) packages work?}

Here we will focus on packages that have been used in behavioral neuroscience, but the general workflow for pose estimation in computer vision research is highly similar. What has made experimentalist-focused toolboxes different is that they provide essential code to generate and train on one's own datasets. Typically, what is available in computer vision focused pose estimation repositories is code to run inference (video analysis) and/or run training of an architecture for specific datasets around which competitions happen (e.g., MS COCO;~\citealp{lin2014microsoft} and MPII pose;~\citealp{andriluka20142d}). While these are two crucial steps, they are not sufficient to develop tailored neural networks for an individual lab or experimentalist. Thus, the ``barrier to entry'' is often quite high to use these tools. It requires knowledge of deep learning languages to build appropriate data loaders, data augmentation pipelines, and training regimes. Therefore, in recent years several packages have not only focused on animal pose estimation networks, but in providing users a full pipeline that allows for (1) labeling a customized dataset (frame selection and labeling tools), (2) generating test/train datasets, (3) data augmentation and loaders, (4) neural architectures, (5) code to evaluate performance, (6) run video inference, and (7) post-processing tools for simple readouts of the acquired machine-labeled data. 

Thus far, around 10 packages have become available in the past 2 years~\cite{mathis2018deeplabcut, pereira2019fast, graving2019fast, pavan2019, arac2019deepbehavior, Zimmermann2020, Bala2020, Liu2020optiflex}. Each has focused on providing slightly different user experiences, modularity, available networks, and balances to the speed/accuracy trade-off for video inference. Several include their (adapted) implementations of the original DeepLabCut or LEAP networks as well~\cite{graving2019fast, Liu2020optiflex}. But the ones we highlight have the full pipeline delineated above as a principle and are open source, i.e., at minimum inference code is available (see Table~\ref{tab:packages}). The progress gained and challenges they set out to address (and some that remain) are reviewed elsewhere~\cite{mathis2020deep, kordingLimitations2019}. Here, we discuss collective aims of these packages (see also Figure~\ref{fig:workflow}).
\medskip

\medskip 
Current packages for animal pose estimation have focused on primarily providing tools to train tailored neural networks to user-defined features.
Because experimentalists need flexibility and are tracking very different animals and features, the most successful packages (in terms of user base as measured by citations and GitHub engagement) are species agnostic. However, given they are all based on advances from prior art in human pose estimation, the accuracy of any one package given the breadth of options that could be deployed (i.e, data augmentation, training schedules, and architectures) will remain largely comparable, if such tools are provided to the user. What will determine performance the most is the input training data provided, and how much capacity the architectures have. 
\medskip

It is notable that using transfer learning has proven to be advantageous for better robustness (i.e., its ability to generalize, see~\citealp{mathis2018deeplabcut, mathis2019TRANSFER, arac2019deepbehavior}), which was first deployed by DeepLabCut (see Table~\ref{tab:packages}). Now, training on large animal-specific datasets has recently been made available in DeepLabCut as well (such as a horse pose dataset with >8,000 annotated images of 30 horses;~\citealp{mathis2019TRANSFER}). This allows the user to bypass the only manual part of curating and labeling ground truth data, and these models can directly be used for inference on novel videos. For DeepLabCut, this is an emerging community-driven effort, with external labs already contributing models and data\footnote{\href{http://modelzoo.deeplabcut.org}{modelzoo.deeplabcut.org}}.
\medskip

\begin{pabox}[label={box:hardware},nameref={Hardware}]{Computing hardware}
\begin{itemize}[leftmargin=*]
\item \textbf{CPU}: The central processing unit (CPU) is the core of a computer and executes  computer programs. 
CPUs work well on sequential or lightly parallelized routines due to the limited number of cores.
\item \textbf{GPU}: A graphical processing unit (GPU) is a specialized computing device designed to rapidly process and alter memory.
GPUs are ideal for computer graphics and often located in graphics cards.
Their highly parallel architecture enables them to be more efficient\footnote{Link to NVIDIA Data Center Deep Learning Product Performance:  \href{https://developer.nvidia.com/deep-learning-performance-training-inference}{developer.nvidia.com/deep-learning-performance-training-inference}} than CPUs for algorithms with many small subroutines which can be launched in parallel.
They can be applied to run DNNs at higher speed~\cite{krizhevsky2012imagenet} and pose estimation in particular~\cite{MathisWarren2018speed,mathis2019TRANSFER, Kane2020dlclive}. 
\item \textbf{Affordability of GPUs}: Modern GPUs are affordable (around 300 - 800 USD for cards than can be used for the pose estimation tools mentioned here; and up to 10,000 USD for high end cards) and ideally suited to run video processing within a single lab in a decentralized way. They can be placed into standard desktop computers, or even ``gaming'' laptops are options. However, to get started it might be easier to test software in cloud computing services first for ease-of-use (i.e. no driver installation). 
\item \textbf{Cloud computing}: Ability to use resources online rapidly (minimal installation) often in a pay-per-use scheme. Two relevant examples are \textit{Google Colaboratory} and \textit{My Binder}. \textit{Google Colaboratory} is an online platform for hosted free GPU use with run times of up to 6 hours. \textit{My Binder} allows turning a Git repository into a collection of interactive notebooks by running them in an executable environment, making your code immediately reproducible by anyone, anywhere (\href{https://mybinder.org/}{mybinder.org}).
\end{itemize}
\end{pabox}

In the future, having the ability to skip labeling and training and run video inference with robust models will lead to more reproducible and scalable research. For example, as we show in other sections of the primer, if the labeling accuracy is not of a high quality, and the data is not diverse enough, then the networks are not able to generalize to so-called ``out-of-domain'' data. If as a community we collectively build stable and robust models that leverage the breadth of behaviors being carried out in laboratories worldwide, we can work towards models that would work in a plug-in-play fashion. We anticipate new datasets and models to become available in the next months to years.
\medskip

All packages, just like all applications of deep learning to video, prefer access to GPU computing resources (See Box~\ref{box:hardware}). On GPUs one experiences faster training and inference times but the code can also be deployed on standard CPUs or laptops. With cloud computing services, such as Google Colaboratory and JupyterLab, many pose estimation packages can simply be deployed on remote GPU resources. This still requires (1) knowledge about these resources, and (2) toolboxes providing so-called ``notebooks'' that can be easily deployed. But, given these platforms have utility beyond just pose estimation, they are worthwhile to learn about.  
\medskip

For the non-GPU aspects, only a few packages have provided easy-to-use graphical user interfaces that allow users with no programming experience to use the tool (see Table~\ref{tab:packages}). Lastly, the available packages vary in their access to 3D tools, multi-animal support, and types of architectures available to the user, which is often a concern for speed and accuracy. Additionally, some packages have limitations on only allowing the same sized videos for training and inference, while others are more flexible. These are all key considerations when deciding which eco-system to invest in learning (as every package has taken a different approach to the API).
\medskip

Perhaps the largest barrier to entry for using deep learning-based pose estimation methods is managing the computing resources (See Box~\ref{box:hardware}, Box~\ref{box:software}). From our experience, installing GPU drivers and the deep learning packages (TensorFlow, PyTorch), that all the packages rely on, is the biggest challenge. To this end, in addition to documentation that is ``user-focused'' (i.e., not just an API for programmers), resources like webinars, video tutorials, workshops, Gitter and community-forums (like StackOverflow and Image Forum SC) have become invaluable resources for the modern neuroscientist. Here, users can ask questions and get assistance from developers and users alike. We believe this has also been a crucial step for the success of DeepLabCut.
\medskip

While some packages provide full GUI-based control over the packages, to utilize more advanced features at least minimal programming knowledge is ideal. Thus, better training for the increasingly computational nature of neuroscience will be crucial. Making programming skills a requirement of graduate training, building better community resources, and leveraging the fast-moving world of technology to harness those computing and user resources will be crucial. In animal pose estimation, while there is certainly an attempt to make many of the packages user-friendly, i.e., to onboard users and have a scalable discussion around common problems, we found user forums to be very valuable~\cite{rueden2019scientific}. Specifically, DeepLabCut is a member of the Scientific Community Image Forum\footnote{\href{https://forum.image.sc/}{forum.image.sc}} alongside other packages that are widely used for image analysis in the life sciences such as Fiji~\cite{schindelin2012fiji}, napari, CellProfiler~\cite{McQuin2018CellProfiler3N} Ilastik~\cite{sommer2011ilastik} and scikit-image~\cite{van2014scikit}. 
\medskip

\begin{pabox}[label={box:software},nameref={Reproducible software architectures for Motion Capture.}]{Reproducible Software}
Often installation of deep learning languages like TensorFlow/Keras \cite{abadi2016tensorflow} and PyTorch \cite{paszke2019pytorch} is the biggest hurdle for getting started. 
\begin{itemize}[leftmargin=*]
\item \textbf{Python virtual environments}: Software often has many dependencies, and they can conflict if multiple versions are required for different needs. Thus, placing dependencies within a contained environment can minimize issues. Common environments include Anaconda (conda) and virtualenv, both for Python code bases. 
\item \textbf{Docker} delivers software in packages called containers, which can be run locally or on servers. Containers are isolated from one another and bundle their own software, libraries and configuration files~(\href{https://www.docker.com/}{docker.com}; \citealp{merkel2014docker}). 
\item \textbf{GitHub}: \href{https://github.com}{github.com} is a platform for developing and hosting software, which uses Git version control. Version control is excellent to have history-dependent versions and discrete workspaces for code development and deployment. GitLab \href{https://gitlab.com/explore}{gitlab.com/explore} also hosts code repositories.

\end{itemize}

\end{pabox}

\section*{Practical considerations for pose estimation (with deep learning)}

\justify As a recent field gaining traction, it is instructive to regard the operability of deep learning-powered pose estimation in light of well-established, often gold standard, techniques.

\subsection*{General considerations and pitfalls}

\justify As discussed in {\it Scope and applications} and as evidenced by the strong adaptation of the tools, deep learning-based pose estimation work well in standard setups with visible animals. The most striking advantage over traditional motion capture systems is the absence of any need for body instrumentation. Although seemingly obvious, the previous statement hides the belated recognition that marker-based motion capture suffers greatly from the wobble of markers placed on the skin surface. That behavior, referred to as ``soft tissue artifact'' among movement scientists and attributable to the deformation of tissues underneath the skin such as contracting muscles or fat, is now known to be the major obstacle to obtaining accurate skeletal kinematics~\footnote{Intra-cortical pins and biplane fluoroscopy give direct, uncontaminated access to joint kinematics. The first, however, is invasive (and entails careful surgical procedures; \citealp{ramsey2003methodological}) whereas the second is only operated in very constrained and complex laboratory settings~\cite{list2017moving}. Both are local to a specific joint, and as such do not strictly address the task of pose estimation.} \cite{camomilla2017}. To make matters worse, contaminated marker trajectories may be harmful in clinical contexts, potentially invalidating injury risk assessment (e.g.~\citealp{smale2017}). Although a multitude of numerical approaches exists to tackle this issue, the most common, yet incomplete, solution is multi-body kinematics optimization (or ``inverse kinematics'' in computer graphics and robotics;~\citealp{begon2018}). This procedure uses a kinematic model and searches for the body pose that minimizes in the least-squares sense the distance between the measured marker locations and the virtual ones from the model while satisfying the constraints imposed by the various joints~\cite{lu1999bone}. Its accuracy is, however, decisively determined by the choice of the underlying model and its fidelity to an individual’s functional anatomy~\cite{begon2018}. In contrast, motion capture with deep learning elegantly circumvents the problem by learning a geometry-aware representation of the body from the data to associate keypoints to limbs~\cite{cao2018openpose,insafutdinov2016deepercut,mathis2020deep}, which, of course, presupposes that one can avoid the ``soft tissue artifact'' when labeling.
\medskip

At present, deep learning-powered pose estimation can be poorly suited to evaluate rotation about a bone's longitudinal axis. From early markerless techniques based on visual hull extraction this is a known problem~\cite{ceseracciu2014comparison}. In marker-based settings, the problem has long been addressed by rather tracking clusters of at least three non-aligned markers to fully reconstruct a rigid segment's six degrees of freedom~\cite{spoor1980rigid}. Performing the equivalent feat in a markerless case is difficult, but it is possible by labeling multiple points (for instance on either side of the wrist to get the lower-limb orientation). Still, recent hybrid, state-of-the-art approaches jointly training under both position and orientation supervision augur very well for video-based 3D joint angle computation~\cite{xu2020eventcap,zhou2020monocular}.
\medskip

With the notable exception of approaches leveraging radio wave signals to predict body poses through walls~\cite{zhao2018through}, deep learning-powered motion capture requires the individuals be visible; this is impractical for kinematic measurements over wide areas. A powerful alternative is offered by Inertial Measurement Units (IMUs)---low-cost and lightweight devices typically recording linear accelerations, angular velocities and the local magnetic field. Raw inertial data can be used for coarse behavior classification across species~\cite{kays2015terrestrial,chakravarty2019novel}. They can also be integrated to track displacement with lower power consumption and higher temporal resolution than GPS~\cite{bidder2015step}, thereby providing a compact and portable way to investigate whole body dynamics (e.g.~\citealp{wilson2018biomechanics}) or, indirectly, energetics~\cite{gleiss2011making}. Recent advances in miniaturization of electronical components now also allow precise quantification of posture in small animals~\cite{pasquet2016wireless}, and open new avenues for kinematic recordings in multiple animals at once at fine motor scales.
\medskip 

\begin{figure*}[h]
    \centering
    \includegraphics[width=.97\textwidth]{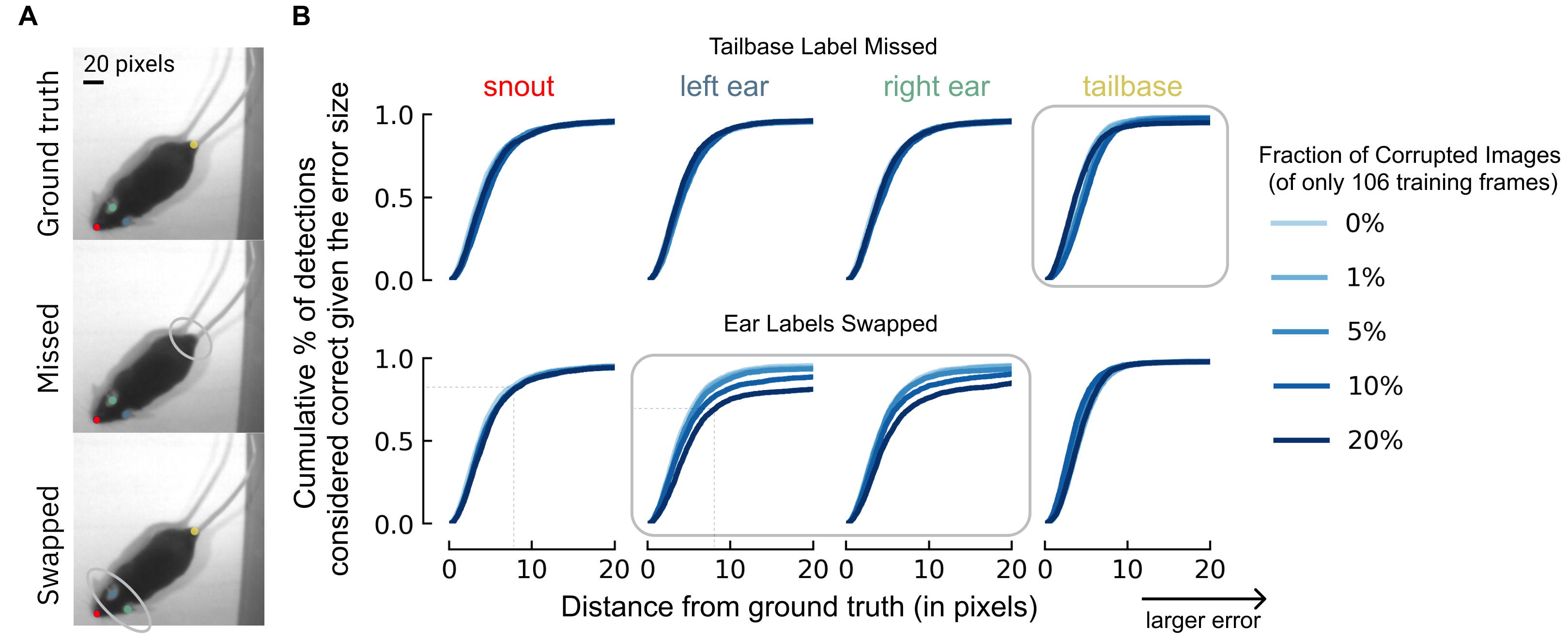}
    \caption{{\bf Labeling Pitfalls: How corruptions affect performance}
   {\bf (A)} Illustration of two types of labeling errors. Top is ground truth, middle is missing a label at the tailbase, and bottom is if the labeler swapped the ear identity (left to right, etc.). {\bf (B)} Using a small dataset of 106 frames, how do the corruptions in A affect the percent of correct keypoints (PCK) as the distance to ground truth increases from 0 pixel (perfect prediction) to 20 pixels (larger error)? The X-axis denotes the difference in the ground truth to the predicted location (RMSE in pixels), whereas Y-axis is the fraction of frames considered accurate (e.g., $\approx$80\% of frames fall within 9 pixels, even on this small training dataset, for points that are not corrupted, whereas for corrupted points this falls to $\approx$65\%). The fraction of the dataset that is corrupted affects this value. Shown is when missing the tailbase label (top) or swapping the ears in $1, 5, 10$ and $20\%$ of frames (of $106$ labeled training images). Swapping vs. missing labels has a more notable adverse effect on network performance.
    }
    \label{fig:corruption}
\end{figure*}

Nonetheless, IMU-based full body pose reconstruction necessitates multiple sensors over the body parts of interest; commercial solutions require up to 17 of them~\cite{roetenberg2009xsens}. That burden was recently eased by utilizing a statistical body model that incorporates anatomical constraints, together with optimizing poses over multiple frames to enforce coherence between the model orientation and IMU recordings---reducing the system down to six sensors while achieving stunning motion tracking~\cite{von2017sparse}. Yet, two additional difficulties remain. The first arises when fusing inertial data in order to estimate a sensor's orientation (for a comprehensive description of mathematical formalism and implementation of common fusion algorithms, see~\citealp{sabatini2011estimating}). The process is susceptible to magnetic disturbances that distort sensor readings and, consequently, orientation estimates~\cite{fan2018magnetic}. The second stems from the necessity to align a sensor's local coordinate system to anatomically meaningful axes, a step crucial (among others) to calculating joint angles (e.g.,~\citealp{lebleu2020lower}). The calibration is ordinarily carried out by having the subject perform a set of predefined movements in sequence, whose execution determines the quality of the procedure. Yet, in some pathological populations (let alone in animals), calibration may be challenging to say the least, deteriorating pose reconstruction accuracy~\cite{vargas2016imu}.
\medskip

A compromise to making the task less arduous is to combine videos and body-worn inertial sensors. Thanks to their complementary nature, incorporating both cues mitigates the limitations of each individual system; i.e., both modalities reinforce one another in that IMUs help disambiguate occlusions, whereas videos provide disturbance-free spatial information~\cite{gilbert2019fusing}. The idea also applies particularly well to the tracking of multiple individuals---even without the use of appearance features, advantageously---by exploiting unique movement signatures contained within inertial signals to track identities over time~\cite{henschel2019simultaneous}.

\subsection*{Pitfalls of using deep learning-based \\motion capture}

\justify Despite being trained on large scale datasets of thousands of individuals, even the best architectures fail to generalize to ``atypical'' postures (with respect to the training set). This is wonderfully illustrated by the errors committed by OpenPose on yoga poses~\cite{huang2019followmeup}. 
\medskip

These domain shifts are major challenges (also illustrated below), and while this is an active area of research with much progress, the easiest way to make sure that the algorithm generalizes well is to label data that is similar to the videos at inference time. However, due to active learning implemented for many packages, users can manually refine the labels on ``outlier'' frames. 
\medskip

Another major caveat of deep learning-powered pose estimation is arguably its intrinsic reliance on high-quality labeled images. This suggests that a labeled dataset that reflects the variability of the behavior should be used. If one -- due to the quality of the video -- cannot reliably identify body parts in still images (i.e., due to massive motion blur, uncertainty about body part (left/right leg crossing) or animal identity) then the video quality should be fixed, or sub-optimal results should be expected. 
\medskip 

To give readers a concrete idea about label errors, augmentation methods, and active learning, we also provide some simple experiments with shared code and data. Code for reproducing these analyses is available at~\href{https://github.com/DeepLabCut/Primer-MotionCapture}{github.com/DeepLabCut/Primer-MotionCapture}.
\medskip 

To illustrate the importance of error-free labeling, we artificially corrupted labels from the trail-tracking dataset from Mathis et al.~\cite{mathis2018deeplabcut}. The corruptions respectively simulate inattentive labeling (e.g., with left–right bodyparts being occasionally confounded), and missing annotation or uncertainty as to whether to label an occluded bodypart. We corrupted $1, 5, 10$ and $20\%$ of the dataset (N=1,066 images) either by swapping two labels or removing one, and trained on $5\%$ of the data. The effect of missing labels is barely noticeable (Figure~\ref{fig:corruption}A). Swapping labels, on the other hand, causes a substantial drop in performance, with an approximate 10\% loss in percentage of correct keypoints (PCK) (Figure~\ref{fig:corruption}B). We therefore reason that careful labeling, more so than labeling a very large number of images, is the safest guard against poor ground truth annotations. We believe that explicitly modeling labeling errors, as done in Johnson and Everingham~\cite{johnson2011learning}, will be an active area of research and integrated in some packages. 
\medskip

\begin{figure*}[b]
    \centering
    \includegraphics[width=.93\textwidth]{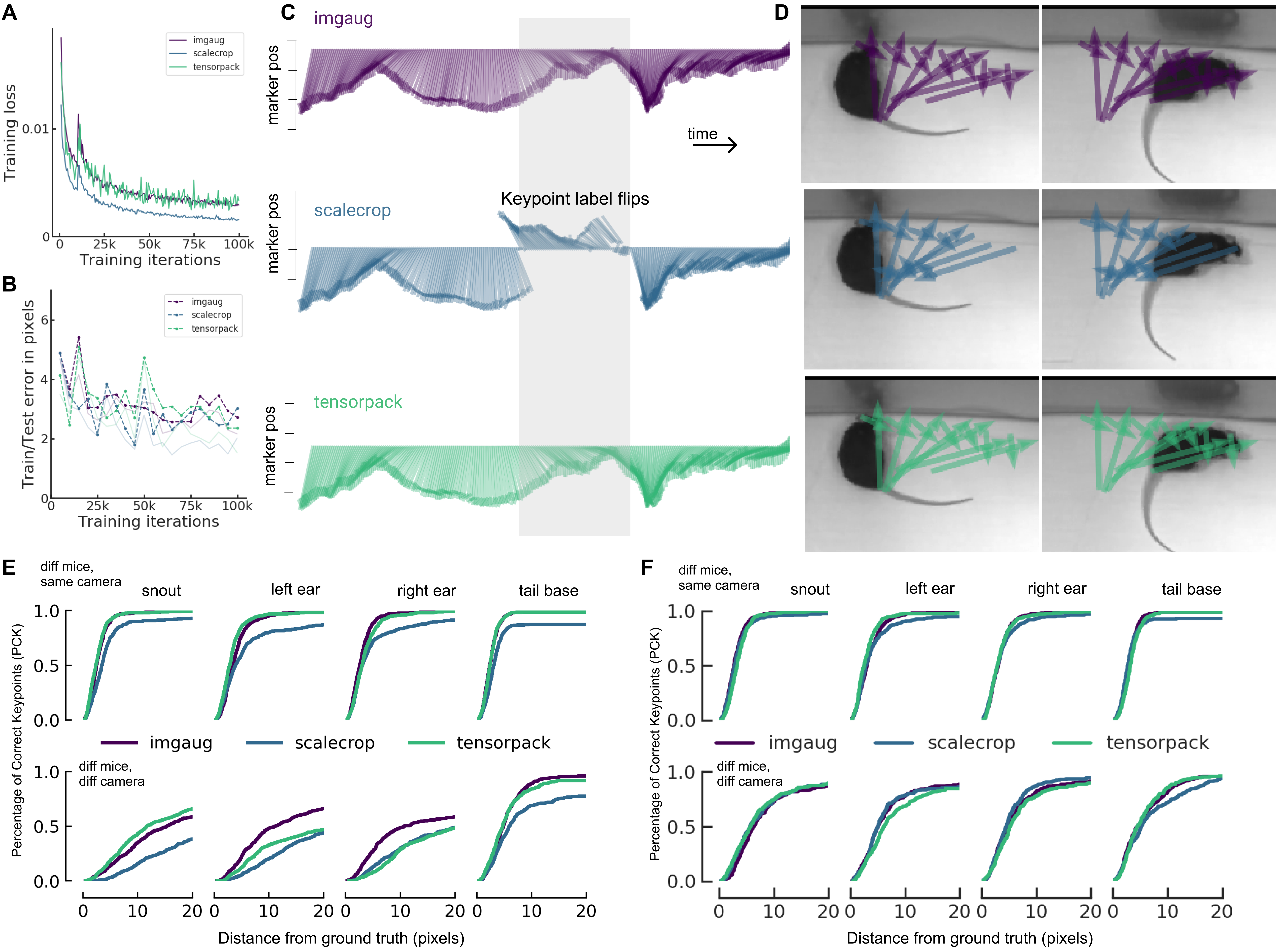}
    \caption{{\bf Data Augmentation Improves Performance}
    Performance of three different augmentation methods on the same dataset of around 100 training images from one short video of one mouse (thus correlated). Scalecrop is configured to only change the scale, and randomly crop images; Imgaug also performs motion blur and rotation ($\pm 180^\circ$) augmentation. Tensorpack performs Gaussian noise and rotation ($\pm 180^\circ$) augmentation. {\bf (A)} Loss over training iterations has plateaued, and {\bf (B)} test errors in pixels appear comparable for all methods. {\bf (C)} Tail base aligned skeletons across time for a video of a different mouse (displayed as a cross connecting snout to tail and left ear to right ear). Note the swap of the ``T'' in the shaded gray zone (and overlaid on the image to the right in {\bf (D)}). Imgaug and tensorpack, which also included full $180^\circ$ rotations, work perfectly). This example highlights that utilizing the rotational symmetry of the data during training can give excellent performance (without additional labeling). {\bf (E)} Performance of the networks on different mice recorded with the same camera (top) and a different camera ($\approx$ 2.5x magnification; bottom). Networks trained with tensorpack and imgaug augmentation generalize much better, and in particular generalize very well to different mice. The generalization to the other camera is difficult, but also works better for  tensorpack and imgaug augmentation. {\bf (F)} Performance of networks on same data as in (E), but after an active learning step, adding $28$ training frames from the higher resolution camera and training for a few thousand iterations. Afterwards, the network generalizes well to both scenarios. 
    }
    \label{fig:AUG2}
\end{figure*}

Even if labeled well, augmentation greatly improves results and should be used. For instance, when training on the example dataset of (highly)-correlated frames from one short video of one individual, the loss nicely plateaus and shows comparable train/test errors for three different augmentation methods (Figure~\ref{fig:AUG2}A, B). The three models also give good performance and generalize to a test video of a different mouse. However, closer inspection reveals that the "scalecrop" augmentation method, which only performs cropping and scaling during training~\cite{nath2019deeplabcut}, leads to swaps in bodyparts with this small training set from only one different mouse (Figure~\ref{fig:AUG2}C, D). The other two methods, which were configured to perform rotations of the training data, could robustly track the posture of the mouse. This discrepancy becomes striking when observing the PCK plots: imgaug and tensorpack outperform scalecrop by a margin of up to $\approx$ 30\% (Figure~\ref{fig:AUG2}E). One simple way to generalize to this additional case is by active learning~\cite{nath2019deeplabcut}, which is also available for some packages. Thereby one annotates additional frames with poor performance (outlier frames) and then trains the network from the final configuration, which thus only requires a few thousand iterations. Adding 28 annotated frames from the higher resolution camera, we get good generalization for test frames from both scenarios (Figure~\ref{fig:AUG2}F). Generally, this illustrates, how the lack of diversity in training data leads to worse performance, but can be fixed by adding frames with poor performance (active learning).

\subsection*{Coping with pitfalls}

\justify Fortunately, dealing with the most common pitfalls is relatively straightforward, and mostly demands caution and common sense. Rules of thumb and practical guidelines are given in Box~\ref{box:pitfalls}. Video quality should be envisaged as a trade-off between storage limitations, labeling precision, and training speed; e.g., the lower the resolution of a video, the smaller the occupied disk space and the faster the training speed, but the harder it gets to consistently identify bodyparts. In practice, DeepLabCut was shown to be very robust to downsizing and video compression, with pose reconstruction degrading only after scaling videos down to a third of their original size or compression by a factor of 1000~\cite{MathisWarren2018speed}.
\medskip

Body parts should be labeled reliably and consistently across frames that preferably capture a variety of behaviors. Note that some packages provide the user means to automatically extract frames differing in visual content based on unsupervised clustering, which simplifies the selection of relevant images in sparse behaviors.
\medskip

Utilize symmetries for training with augmentation and try to include image augmentations that are helpful. Use the strongest model (given the speed requirements). Check performance and actively grow the training set if errors are found. 
\medskip 

\begin{pabox}[label={box:pitfalls},nameref={Pity}]{Avoiding pitfalls.}
\begin{itemize}[leftmargin=*]
\item \textbf{Video quality}: While deep learning based methods are more robust than other methods, and can even learn from blurry, low-resolution images, you will make your life easier by recording quality videos. 
\item \textbf{Labeling}: Label accurately and use enough data from different videos. 10 videos with 20 frames each is better than 1 video with 200 frames. Check labeling quality. If multiple people label, agree on conventions - i.e. be sure that for a larger body part (like back of mouse) the same location is labeled.
\item \textbf{Dataset curation}: Collect annotation data from the full repertoire of behavior (different individuals, backgrounds, postures). Automatic methods of frame extraction exist, but the videos need to be manually selected.
\item \textbf{Data Augmentation}: Are there specific features you know happen in your videos, like motion blur or contrast changes? Can rotational symmetry, or mirroring be exploited? Then use an augmentation scheme that can build this into training. 
\item \textbf{Optimization}: Train until loss plateaus, and do not over-train. Check that it worked by looking at performance on training images (both quantitatively and visually), ideally across ``snapshots" (i.e. train iterations of the network). If that works, look at test images. Does the network generalize well? Note that, even if everything is proper, train and test performance can be different due to over-fitting on idiosyncrasies of training set.  Bear in mind that the latest iterations may not be the ones yielding the smallest errors on the test set. It is therefore recommended to store and evaluate multiple snapshots.
\item \textbf{Cross-validation}: You can compare different parameters (networks, augmentation, optimization) to get the best performance (see Figure~\ref{fig:corruption}).
\end{itemize}
\end{pabox}

Pose estimation algorithms can make different types of errors: jitter, inversion (e.g. left/right), swap (e.g. associating body part to another individual) and miss~\cite{ruggero2017benchmarking}. Depending on the type of errors, different causes need to be addressed (i.e., check the data quality for any human-applied mistakes~\cite{mathis2018deeplabcut}, use suitable augmentation methods). Also for some cases, post processing filters can be useful (such as Kalman filters), but also graphical models or other methods that learn the geometry of the bodyparts. We also believe that future work will explicitly model labeling errors during training.

\section*{What to do with motion capture data?}

Pose estimation with deep learning is to relieve the user of the painfully slow digitization of keypoints. With markerless tracking you need to annotate a much smaller dataset and this can be applied to new videos. Pose estimation also serves as a springboard to a plethora of other techniques. Indeed, many new tools are specifically being developed to aid users of pose estimation packages to analyze movement and behavioral outputs in a high-throughput manner. Plus, many such packages existed pre-deep learning and can now be leveraged with this new technology as well. While the general topic of what to do with the data is beyond this primer, we will provide a number of pointers. These tools fall into three classes: time series analysis, supervised, and unsupervised learning tools.
\medskip

A natural step ahead is the quantitative analysis of the keypoint trajectories. The computation of linear and angular displacements, as well as their time derivatives, lays the ground for detailed motor performance evaluation---a great introduction to elementary kinematics can be found in~\cite{Winter2009}, and a thorough description of 151 common metrics is given in~\cite{schwarz2019systematic}. These have a broad range of applications, of which we highlight a system for assessing >30 behaviors in groups of mice in an automated way~\cite{de2019real}, or an investigation of the evolution of gait invariants across animals~\cite{catavitello2018kinematic}. Furthermore, kinematic metrics are the basis from which to deconstruct complex whole-body movements into interpretable motor primitives, non-invasively probing neuromuscular control~\cite{longo2019biomechanics}. Unsupervised methods such as clustering methods~\cite{Pedregosa2011}, MotionMapper~\cite{Berman2014},  MoSeq~\citep{wiltschko2015mapping}, or variational autoencoders~\cite{luxem2020identifying} allow the extraction of common ``kinematic behaviors'' such as turning, running, rearing. Supervised methods allow the prediction of human defined labels such as ``attack'' or ```freezing.'' For this, general purpose tools such as scikit-learn~\cite{Pedregosa2011} can be ideal, or tailored solutions with integrated GUIs such as JAABA can be used~\citep{kabra2013jaaba}. Sturman et al. have developed an open source package to utilize motion capture outputs together with classifiers to automate human annotations for various behavioral tests (open field, elevated plus maze, forced swim test). They showed that these open source methods outperform commercially available platforms~\cite{sturman2020deep}.  
\medskip

Kinematic analysis, together with simple principles derived from physics, also allows the calculation of the energy required to move about, a methodology relevant to understanding the mechanical determinants of the metabolic cost of locomotion (e.g.~\citealp{saibene2003biomechanical}) or informing the design of bio-inspired robots (e.g.~\citealp{li2017mechanical,nyakatura2019reverse}).

\subsection*{Modeling and motion understanding}

\justify Looking forward, we also expect that the motion capture data will be used to learn task-driven and data-driven models of the sensorimotor as well as the motor pathway. We have recently provided a blueprint combining human movement data, inverse kinematics, biomechanical modeling and deep learning~\cite{sandbrink2020task}. Given the complexity of movement, as well as the highly nonlinear nature of the sensorimotor processing~\cite{madhav2020synergy, nyakatura2019reverse}, we believe that such approaches will be fruitful to leverage motion capture data to gain insight into brain function. 

\section*{Perspectives}

As we highlighted thus far in this primer, markerless motion capture has reached a mature state in only a few years due to the many advances in machine learning and computer vision. While there are still some challenges left~\cite{mathis2020deep}, this is an active area of research and advances in training schemes (such as semi-supervised and self-supervised learning) and model architectures will provide further advances and even less required manual labour. Essentially, now every lab can train appropriate algorithms for their application and turn videos into accurate measurements of posture. If setups are sufficiently standardized, these algorithms already broadly generalize, even across multiple laboratories as in the case of the International Brain Lab~\cite{Harris2020dataIBL}. But how do we get there, and how do we make sure the needs of animal pose estimation for neuroscience applications are met?

\subsection*{Recent developments in deep learning}

\justify  Innovations in the field of object recognition and detection affect all aforementioned parts of the algorithm, as we discussed already in the context of using pre-trained representations. An emerging relevant research direction in machine learning is large scale semi-supervised and self-supervised representation learning (SSL). In SSL, the problem of pre-training representations is no longer dependent on large labeled datasets, as introduced above. Instead, even larger databases comprised of unlabeled examples---often multiple orders of magnitude larger than the counterparts used in supervised learning---can be leveraged.
A variety of SSL algorithms are becoming increasingly popular in all areas of machine learning.
Recently, representations obtained by large-scale self-supervised pre-training began to approach or even surpass performance of the best supervised methods.
Various SSL methods \cite{oord2018representation, logeswaran2018efficient,wu2018unsupervised,henaff2019data,tian2019contrastive,hjelm2018learning,bachman2019learning,he2019momentum,chen2020simple, wu2018unsupervised,hjelm2018learning,bachman2019learning,he2019momentum,chen2020simple} made strides in both image recognition \cite{chen2020simple}, speech processing \citep{schneider2019wav2vec,baevski2019vq,baevski2020wav2vec,ravanelli2020multi} and NLP~\cite{devlin2019bert,Liu2019roberta}, already starting to outperform models obtained by supervised pre-training on large datasets. Considering that recent SSL models for computer vision are continued to being shared openly (e.g.~\citealp{xie2020noisy,chen2020simple}), it can be expected to impact and improve new model development in pose estimation, especially if merely replacing the backend model is required.
On top, SSL methods can be leveraged in end-to-end models for estimating keypoints and poses directly from raw, unlabeled video \cite{umer2020self, tung2017self, kocabas2019self}.
Approaches based on graph neural networks \cite{scarselli2008graph}  can encode priors about the observed structure and model correlations between individual keypoints and across time \cite{cai2019exploiting}. For some applications (like modeling soft tissue or volume) full surface reconstructions are needed and this area has seen tremendous progress in recent years~\cite{guler2018densepose,sanakoyeu2020transferring, Zuffi2019ICCV}. Such advances can be closely watched and incorporated in neuroscience, but we also believe our field (neuroscience) is ready to innovate in this domain too.

\subsection*{Pose estimation specifically for neuroscience}

\justify The goal of human pose estimation---aside from the purely scientific advances for object detection---range from person localization in videos, self-driving cars and pedestrian safety, to socially aware AI, is related to, but does differ from, the applied goals of animal pose estimation in neuroscience. Here, we want tools that give us the highest precision, with the most rapid feedback options possible, and we want to train on small datasets but have them generalize well. This is a tall order, but so far we have seen that the glass is (arguably more than) half full. How do we meet these goals going forward? While much research is still required, there are essentially two ways forward: datasets and associated benchmarks, and algorithms.

\subsection*{Neuroscience needs (more) benchmarks}

\justify In order to push the field towards innovations in areas the community finds important, setting up benchmark datasets and tasks will be crucial (i.e., the Animal version of ImageNet). The community can work towards sharing and collecting data of relevant tasks and curating it into benchmarks. This also has the opportunity of shifting the focus in computer vision research: Instead of ``only'' doing human pose estimation, researchers probably will start evaluating on datasets directly relevant to neuroscience community. Indeed there has been a recent interest in more animal-related work at top machine learning conferences~\cite{khan2020animalweb, sanakoyeu2020transferring}, and providing proper benchmarks for such approaches would be ideal. 
\medskip

For animals, such efforts are developing: Khan et al. recently shared a dataset comprising  22.4K annotated faces from 350 diverse species~\cite{khan2020animalweb} and Labuguen announced a dataset of 13K annotated macaque~\cite{labuguen2020macaquepose}.
We recently released two benchmark datasets that can be evaluated for state-of-the-art performance~\footnote{\href{https://paperswithcode.com}{paperswithcode.com}} on within domain and out-of-domain data~\footnote{\href{http://horse10.deeplabcut.org}{horse10.deeplabcut.org}}. The motivation is to train on a limited number of individuals and test on held out animals (the so-called ``out-of-domain'' issue)~\cite{mathis2019TRANSFER, mathisimagenet2020}. We picked horses due to the variation in coat colors (and provide >8K labeled frames). Secondly, to directly study the inherent shift in domain between individuals, we set up a benchmark for common image corruptions, as introduced by Hendrycks et al.~\cite{Hendrycks2019} that uses the image corruptions library proposed by Michaelis et al.~\cite{michaelis2019dragon}. 
\medskip

Of course these aforementioned benchmarks are not sufficient to cover all the needs of the community, so we encourage consortium-style efforts to also curate data and provide additional benchmarks. Plus, making robust networks is still a major challenge, even when trained with large amounts of data~\cite{beery2018recognition, geirhos2020shortcut}. In order to make this a possibility it will be important to develop and share common keypoint estimation benchmarks for animals as well as expand the human ones to applications of interest, such as sports~\cite{huang2019followmeup}.

\subsection*{Sharing Pre-trained Models}

\justify We believe another major step forward will be sharing pre-trained pose estimation networks.  If as a field we were to annotate sufficiently diverse data, we could train more robust networks that broadly generalize. This success is promised by other large scale data sets such as MS COCO~\cite{lin2014microsoft} and MPII pose~\cite{andriluka20142d}.
In the computer vision community, sharing model weights such that models do not need to be retrained has been critical for progress. For example, the ability to download pre-trained ImageNet weights is invaluable---training ImageNet from scratch on a standard GPU can take more than a week. Now, they are downloaded within a few seconds and fine tuned in packages like DeepLabCut.
However even for custom training setups, sharing of code and easy access to cloud computing resources enables smaller labs to train and deploy models without investment in additional lab resources.
Pre-training a typical object recognition model on the ILSVC is now possible on the order of minutes for less than 100 USD \cite{coleman2017dawnbench} thanks to high-end cloud computing, which is also feasible for labs lacking the necessary on-site infrastructure (Box~\ref{box:hardware}).
\medskip

In neuroscience, we should aim to fine tune even those models; namely, sharing of mouse-specific, primate-specific weights will drive interest and momentum from researchers without access to such data, and further drive innovations. Currently, only DeepLabCut provides model weights (albeit not at the time of the original publication) as part of the recently launched Model Zoo (\href{http://modelzoo.deeplabcut.org/}{modelzoo.deeplabcut.org}). Currently it contains models trained on MPII pose~\cite{insafutdinov2016deepercut}, dog and cat models as well as contributed models for primate facial recognition, primate full body recognition~\cite{labuguen2020macaquepose} and mouse pupil detection (Figure~\ref{fig:workflow}). Researchers can also contribute in a citizen-science fashion by labeling data on the web (\href{http://contrib.deeplabcut.org}{contrib.deeplabcut.org}) or by submitting models. 
\medskip

Both datasets and models will benefit from common formatting to ease sharing and testing. Candidate formats are HDF5 (also chosen by NeuroData Without Borders~\cite{teeters2015neurodata} and DeepLabCut), TensorFlow data\footnote{%
\href{https://www.tensorflow.org/api_docs/python/tf/data}{tensorflow.org/api\_docs/python/tf/data}},
and/or PyTorch data\footnote{%
\href{https://pytorch.org/docs/stable/torchvision/datasets.html}{pytorch.org/docs/stable/torchvision/datasets.html}}. Specifically, for models, proto-buffer formats for weights are useful and easy to share~\cite{Kane2020dlclive, lopes2015bonsai} for deployment to other systems.
Platforms such as OSF and Zenodo allow banking of weights, and some papers (e.g.~\citealp{barrett2020manual, sturman2020deep}) have also shared their trained models. We envision that having easy-to-use interfaces to such models will be possible in the future.

\medskip

These pre-trained pose estimation networks hold several promises: it saves time and energy (as different labs do not need to annotate and train networks), as well as contributes to reproducibility in science. Like many other forms of biological data, such as genome sequences, functional imaging data, behavioral data is notoriously hard to analyze in standardized ways. Lack of agreement can lead to different results, as pointed out by a recent landmark study comparing the results achieved by 70 independent researchers analyzing nine hypothesis in shared imaging data~\cite{botvinik2020variability}. To increase reproducibility in behavioral science, video is a great tool~\cite{gilmore2017video}. Analyzing behavioral data is complex, owing to its unstructured, large-scale nature, which highlights the importance of shared analysis pipelines. Thus, building robust architectures that extract the same behavioral measurements in different laboratories would be a major step forward.

\section*{Conclusions}

Deep learning based markerless pose estimation has been broadly and rapidly adopted in the past two years. This impact was, in part, fueled by open-source code: by developing and sharing packages in public repositories on GitHub they could be easily accessed for free and at scale. These packages are built on advances (and code) in computer vision and AI, which has a strong open science culture. Neuroscience also has strong and growing open science culture~\cite{white2019future}, which greatly impacts the field as evidenced by tools from the Allen Institute, the UCLA Miniscope~\cite{aharoni2019all}, OpenEphys~\cite{siegle2017open}, and Bonsai~\cite{lopes2015bonsai} (just to name a few). 
\medskip

Moreover, Neuroscience and AI have a long history of influencing each other~\cite{hassabis2017neuroscience}, and research in Neuroscience will likely contribute to making AI more robust~\cite{SINZ2019, hassabis2017neuroscience}. The analysis of animal motion is a highly interdisciplinary field at the intersection of biomechanics, computer vision, medicine and robotics with a long tradition~\cite{klette2008understanding}. The recent advances in deep learning have greatly simplified the measurement of animal behavior, which, as we and others believe~\cite{krakauer2017neuroscience}, in turn will greatly advance our understanding of the brain. 

\begin{flushleft}
\textbf{Acknowledgments:} 
\end{flushleft}
We thank Yash Sharma for discussions around future directions in self-supervised learning, Erin Diel, Maxime Vidal, Claudio Michaelis, Thomas Biasi for comments on the manuscript.
Funding was provided by the Rowland Institute at Harvard 
University (MWM, AM),
the Chan Zuckerberg Initiative (MWM, AM, JL)
and the German Federal Ministry of Education and Research (BMBF) through the Tübingen AI Center (StS; FKZ: 01IS18039A).
StS thanks the International Max Planck Research School for Intelligent Systems (IMPRS-IS) and acknowledges his membership in the European Laboratory for Learning \& Intelligent Systems (ELLIS) PhD program.
The authors declare no conflicts of interest. M.W.M. dedicates this work to Adam E. Max.

\section*{References}


\end{document}